\documentclass[twoside,11pt]{article}

\usepackage{jmlr2e}
\usepackage{enumitem}
\usepackage{soul}

\hypersetup{citecolor=MidnightBlue}
\hypersetup{urlcolor=MidnightBlue}
\hypersetup{linkcolor=red}

\usepackage{amsmath}
\usepackage{amsfonts}
\usepackage{graphicx}
\usepackage{natbib}
\usepackage{multirow}
\usepackage{enumitem}

\usepackage{url} 

\usepackage[usenames,dvipsnames,table]{xcolor}

\usepackage{cleveref}
\usepackage{amssymb}
\usepackage{bbm}
\usepackage{tikz}
\usetikzlibrary{patterns}

\DeclareMathOperator*{\argmax}{arg\hspace{0.1ex}max}
\DeclareMathOperator*{\argmin}{arg\hspace{0.1ex}min}

\usepackage{tikz}
\usepackage{tkz-euclide}
\usepackage{mdframed}

\usepackage{pifont}
\newcommand{\cmark}{\ding{51}}%
\newcommand{\xmark}{\ding{55}}%

\newcommand{\KL}{\text{KL}}
\newcommand{\SM}{\text{S}}
\newcommand{\Ra}{\text{D}_\alpha}
\newcommand{\bz}{{\bf z}}

\newcommand{\bs}{{\bf s}}
\newcommand{\bt}{{\bf t}}
\newcommand{\bsf}{{\bf \mathfrak s}}
\newcommand{\bmu}{\boldsymbol \mu}
\newcommand{\bnu}{\boldsymbol \nu}
\newcommand{\bsigma}{\boldsymbol \Sigma}
\newcommand{\bpsi}{\boldsymbol \Psi}
\newcommand{\bphi}{\boldsymbol \Phi}
\newcommand{\bxi}{\boldsymbol \xi}
\newcommand{\bC}{{\bf C}}
\newcommand{\bR}{{\bf R}}

\newcommand{\bH}{{\bf H}}
\newcommand{\bU}{{\bf U}}

\definecolor{gg1}{HTML}{B2182B}
\definecolor{gg2}{HTML}{EF8A62}
\definecolor{gg3}{HTML}{FDDBC7}
\definecolor{gg4}{HTML}{D1E5F0}
\definecolor{gg5}{HTML}{67A9CF}

\definecolor{lightgray}{RGB}{200, 200, 200}

\usepackage{lastpage}
\jmlrheading{26}{2025}{1-\pageref{LastPage}}{6/24; Revised 5/25}{9/25}{24-0878}{Charles C. Margossian, Loucas Pillaud-Vivien and Lawrence K. Saul}

\ShortHeadings{Variational Inference for Uncertainty Quantification}{Margossian, Pillaud-Vivien and Saul}
\firstpageno{1}

\begin{document}

\title{Variational Inference for Uncertainty Quantification:\\ an Analysis of Trade-offs}

\author{\name Charles C. Margossian \email charles.margossian@ubc.ca \\
       \addr Department of Statistics \\
        University of British Columbia\\
       Vancouver, BC, Canada
       \AND
       \name Loucas Pillaud-Vivien \email loucas.pillaud-vivien@enpc.fr \\
       \addr CERMICS Laboratory\\
       Ecole des Ponts ParisTech \\
       Champs-sur-Marne, France
       \AND
       \name Lawrence K. Saul \email lsaul@flatironinstitute.org \\
       \addr  Center for Computational Mathematics\\
       Flatiron Institute \\
       New York, NY, USA}

\editor{Barbara Engelhardt}

\maketitle

\bigskip
\begin{abstract} %
Given an intractable distribution $p$, the problem of variational inference (VI) is to find the best approximation from some more tractable family $\mathcal Q$.
Commonly, one chooses $\mathcal{Q}$ to be a family of factorized distributions (i.e., the mean-field assumption), even though~$p$ itself does not factorize.
We show that this mismatch can lead to an \textit{impossibility theorem}: if $p$ does not factorize and furthermore has a non-diagonal covariance matrix, then any factorized approximation $q\!\in\!\mathcal{Q}$ can correctly estimate at most one of the following three measures of uncertainty: (i) the marginal variances, (ii) the marginal precisions, or (iii) the generalized variance (which for elliptical distributions is closely related to the entropy). In practice, the best variational approximation in $\mathcal{Q}$ is found by minimizing some divergence $D(q,p)$ between distributions, 
and so we ask: how does the choice of divergence determine which measure of uncertainty, if any, is correctly estimated by VI?
We consider the classic Kullback-Leibler divergences, the more general $\alpha$-divergences, and a score-based divergence which compares $\nabla \log p$ and $\nabla \log q$.
We thoroughly analyze the case where $p$ is a Gaussian and $q$ is a (factorized) Gaussian.
In this setting, we show that all the considered divergences can be \textit{ordered} based on the estimates of uncertainty they yield as objective functions for~VI. 
Finally, we empirically evaluate the validity of this ordering when the target distribution $p$ is not Gaussian.

\end{abstract}

\ \\[-0.5pt]


\section{Introduction}

In many problems, it is useful to approximate an intractable distribution, $p$, by a more tractable distribution, $q$. This problem can be posed as an optimization, one whose goal is to discover the best-matching distribution $q$ within some larger, parameterized family~$\mathcal{Q}$ of approximating distributions. Optimizations of this sort are the subject of a large literature on variational inference \citep[VI;][]{Jordan:1999, Wainwright:2008, Blei:2017}. 

Formally, suppose we wish to approximate a \textit{target} distribution $p$ over latent variables \mbox{$\bz = z_{1:n} \in \mathcal Z$},
and let $\mathcal Q$ be the family of approximating distributions. VI attempts to solve an optimization of the form
\begin{equation}
  q^* = \underset{q \in \mathcal Q}{\text {argmin}} \ D(q, p),
\label{eq:argminDqp}
\end{equation}
where $D$ is a \textit{divergence} satisfying (i) $D(q, p) \ge 0$ for all $q\in\mathcal{Q}$ and (ii) $D(q, p) = 0$ if and only if $p = q$.
In most applications, $\mathcal Q$ is not sufficiently rich to capture all the features of~$p$---that is $p \notin \mathcal Q$---and so, even after minimizing $D(q, p)$, the optimal approximation $q$ must be compromised in some way.
A crucial question is then whether $q$ can still capture the features of $p$ that are important for a given application, even if other (less relevant) features are poorly estimated.
In this paper, we study certain trade-offs that arise from the
choice of $\mathcal Q$, 
and we analyze how these trade-offs are resolved by the particular choice of divergence in \cref{eq:argminDqp}.



Perhaps the most common choice for $\mathcal Q$ is the family of \textit{factorized} approximations,
\begin{equation}
  q(\bz) = \prod_{i = 1}^n q_i(z_i).
  \label{eq:q_factor}
\end{equation}
We refer to this setting as F-VI. Variational families of this form originated in the mean-field approximation for physical models of ferromagnetism~\citep{Parisi:1988}.
Factorized approximations are computationally convenient for many applications of machine learning because the number of variational parameters scales linearly with the dimensionality of $\bz$ \citep[e.g][]{Bishop:2002, Blei:2012, Giordano:2023}; in addition 
better scaling can be achieved for large data sets 
via amortization, which further restricts the form of $\mathcal Q$~\citep{Kingma:2014, Margossian:2024}.

Our paper begins by formalizing the inherent trade-offs that arise in F-VI. Consider a target $p$ which does \textit{not} factorize and has a finite non-diagonal covariance matrix~$\bsigma$.
(The latter implies that some of the components of $\bz$ are \textit{correlated}, which is stronger than dependence.)
By its very nature, the factorized approximation in eq.~(\ref{eq:q_factor}) cannot estimate the correlations between different components of $\bz$.
Nonetheless, a factorized approximation may still capture useful features of $p$, such as its mean and some quantification of uncertainty.
%
%
There are many ways to characterize the uncertainty of a multivariate distribution~$p$. In this paper, we specifically consider:
\begin{enumerate}
    \item[(i)] the marginal variances, $\Sigma_{ii}$, equal to the diagonal elements of the covariance matrix, \\[-4ex]
    \item[(ii)] the marginal precisions, $\Sigma^{-1}_{ii}$, 
     equal to the diagonal elements\footnote{Throughout the paper, for any square matrix $\mathbf{A}$, we use $A^p_{ij}$  to denote $(A^p)_{ij}$, the $ij^\text{th}$ element of the~$p^\text{th}$ matrix power of $A$, and we use $(A_{ij})^p$ to denote the $p^\text{th}$ scalar power of the matrix element~$A_{ij}$.} of the precision matrix (i.e., the inverse covariance matrix), and \\[-4ex]
    \item[(iii)] the generalized variance, $|\bsigma|$, defined as the determinant of the covariance matrix.
\end{enumerate}
Which measure of uncertainty to pursue depends on the application.
It is especially common in Bayesian statistics to report the marginal variances of interpretable variables~\citep{Gelman:2013}, and in some models, the marginal precisions~\citep{Bernado:2010}.
On the other hand, sometimes it is useful to report a single scalar measure of uncertainty; this is provided by the generalized variance, a measure of uncertainty that for elliptical distributions is closely related to the differential
entropy in 
information theory~\citep{Mackay:2003}.

\begin{table}
  \begin{center}
 \addtolength{\tabcolsep}{1ex}
     \begin{tabular}{rrl} 
     \rowcolor{Cerulean!10} {\bf Divergence} & \hspace{2ex}{\bf Notation } & {\bf Definition}  \\
    {(Reverse) Kullback-Leibler} & $\KL(q || p)$ & $\mathbb E_q\!\left[\log\frac{q}{p}\right]$ \\
      \rowcolor{Cerulean!10} 
    {(Forward) Kullback-Leibler} & $\KL(p || q)$ & $\mathbb E_p\!\left[\log\frac{p}{q}\right]$ \\
    $\alpha$-family with $\alpha \in \mathbb  R^+ \backslash \{0, 1\} $ & $\Ra (p || q)$ & {$\frac{1}{\alpha(\alpha - 1)}\, \mathbb{E}_q\! 
    \left[\frac{p^\alpha}{q^\alpha} - 1 \right]$} \\
    \rowcolor{Cerulean!10} 
    (Reverse) Score-based & $\SM (q || p)$ & $\mathbb{E}_q\!\left[\big\|\nabla \log \frac{q}{p}\big\|^2_{\text{Cov}(q)}\right]$ \\
     (Forward) Score-based & $\SM (p || q)$ & $\mathbb{E}_p\!\left[\big\|\nabla \log \frac{q}{p}\big\|^2_{\text{Cov}(p)}\right]$  
    \end{tabular}
 \addtolength{\tabcolsep}{-1ex}  
    \caption{{Divergences that we analyze in this paper for FG-VI. The score-based divergences~\citep{Cai:2024} are defined here for the special case that the base distribution is Gaussian.}}
    \label{tab:divergences}
  \end{center}
\end{table}

The first main result of this paper is an impossibility theorem: it states that any factorized approximation, of the form in eq.~(\ref{eq:q_factor}), can match \textit{at most one} of the above three measures of uncertainty.
An immediate consequence is that there exists no universally ``best'' factorized approximation of $p$; what is best for one application may be of little use for another.

This ambiguity is reflected by 
the fact that different divergences return different solutions to eq.~(\ref{eq:argminDqp}).
This, in turn, begs the following question: which divergence is best suited for a particular application?
By far the most popular choice in practice is the Kullback-Leibler divergence, $\KL(q || p)$, which measures the disagreement between $\log p$ and $\log q$.
Other valid choices include the ``forward'' $\KL(p || q)$ \citep[e.g.,][]{Naesseth:2020, Vehtari:2020}, the $\alpha$-divergences $\Ra(p||q)$ \citep{Minka:2005, Li:2016} which interpolate between $\KL(q||p)$ and $\KL(p||q)$ when $\alpha\!\in\!(0,1)$, and score-based (or Fisher) divergences \citep{Courtade:2016, Cai:2024, cai2024b, Modi:2025}, which measure the discrepancy between $\nabla \log p$ and $\nabla \log q$. 
Table~\ref{tab:divergences} lists the divergences that we consider in this paper.

We investigate how each divergence in Table~\ref{tab:divergences} resolves the constraints imposed by the impossibility theorem. To make progress, we identify a particular setting of VI where these questions can be fully analyzed. This is the setting where~$q$ belongs to a the family of Gaussians with a diagonal covariance matrix---a common choice in black box VI \citep{Kucukelbir:2017}, which we call FG-VI---and where $p$ is a Gaussian with a non-diagonal covariance matrix:
\begin{align}
    p &= \text{normal}(\bmu, \bsigma), \label{eq:p_normal} \\
    q &= \text{normal}(\bnu, \bpsi), \label{eq:q_normal}
\end{align}
with $\Psi_{ij}\!=\!0$ whenever $i\!\neq\!j$. This setting, though relatively simple, is already rich enough to provide several counter-intuitive results, and it reveals in a stark way that different inferential goals are best served by different divergences.

Our work in this setting builds on earlier analyses of F-VI based on minimizing KL-divergences~\citep{Margossian:2023}. In particular, we also analyze F-VI when~$q$ is chosen to minimize the $\alpha$-divergences or score-based divergences in Table~\ref{tab:divergences}; this requires more technical machinery because there do not exist closed-form solutions (even in this simple setting) for the variational approximation that minimizes these divergences. Thus we fill a gap in the VI literature for these non-classical divergences by further characterizing their variational approximations.




The second main result of the paper is an \textit{ordering} of all the divergences in Table~\ref{tab:divergences} based on the estimates they yield of the marginal variances, marginal precisions, and generalized variance when $p$ and $q$ are Gaussian.
This ordering provides a natural and interpretable way of comparing divergences based on a downstream inferential task.
While there exist many ways to compare divergences \citep[e.g.,][]{Gibbs:2002}, we believe that this ordering should be
particularly helpful to practitioners.
The ordering not only illuminates
which divergence may yield the best estimate of some measure of uncertainty; it also cautions us that certain divergences can fail spectacularly---for instance, producing solutions for F-VI with variances that collapse to 0 or blow up to infinity. \\[0.1pt]

\noindent 
\fcolorbox{Cerulean!10}{Cerulean!10}{\textit{Outline.}} In section~\ref{sec:summary}, we state the main results of the paper---the impossibility theorem, and the ordering theorem---and we show how and where they fit into the existing literature. In section~\ref{sec:impossibility}, we prove the impossibility theorem by analyzing the pairwise trade-offs that arise in F-VI between different measures of uncertainty. In section~\ref{sec:divergence}, we calculate the divergences in Table~\ref{tab:divergences} and characterize the solutions to eq.~(\ref{eq:argminDqp}) when $p$ and $q$ are Gaussian, as in eqs.~(\ref{eq:p_normal}--\ref{eq:q_normal}). In section~\ref{sec:ordering}, we prove the ordering theorem by comparing the estimates of uncertainty obtained by minimizing these divergences. In section~\ref{sec:experiments}, we empirically investigate the validity of this ordering, on a variety of targets and data sets, when $q$ is Gaussian but~$p$ is not. Finally, in section~\ref{sec:discuss}, we discuss our results in light of various applications which require uncertainty quantification (e.g., Bayesian statistics, pre-conditioning of Markov chain Monte Carlo, and information theory), and we explore directions for future work.

\section{Summary of Contributions}
\label{sec:summary}

The main results of the paper are an impossibility theorem for F-VI, starting from eq.~(\ref{eq:q_factor}), and an ordering theorem for the divergences in Table~\ref{tab:divergences}, starting from eqs.~(\ref{eq:p_normal}--\ref{eq:q_normal}). 

\subsection{Main Results}

%
%
We begin with the impossibility theorem.
To state the result, the following notation is useful.
For any square matrix, we use $\text{diag}(\cdot)$ to denote the vector formed from its diagonal elements, and we use vector inequalities, such as $\text{diag}(\bpsi)\!\leq\!\text{diag}(\bsigma)$, to denote that $\Psi_{ii}\!\leq\!\Sigma_{ii}$ for all~$i$.
Furthermore, we use denote $A^{-1}_{ij}$ to denote the $(ij)^\text{th}$ element of ${\bf A}^{\!-1}$.
Note that for a non-diagonal matrix, $A^{-1}_{ij} \neq 1 / A_{ij}$.

\begin{mdframed}[hidealllines = true, backgroundcolor = Cerulean!10]
\begin{theorem}[Impossibility theorem for F-VI]  \label{thm:impossibility}
  Let $p$ and $q$ be distributions with covariances $\bsigma$ and $\bpsi$, respectively, where $\bpsi$ is diagonal but $\bsigma$ is not.~Then
  \begin{enumerate}
    \item (Variance matching) \\
    If $\text{diag}({\bpsi})\!=\!\text{diag}({\bsigma})$, then $|\bpsi|\! > \!|\bsigma|$ and $\text{diag}(\bpsi^{-1})\!\leq\!\text{diag}(\bsigma^{-1})$, and this last inequality is strict for at least one component along the diagonal.
    \item (Precision matching) \\
     If $\text{diag}({\bpsi}^{-1})\!=\!\text{diag}({\bsigma^{-1}})$,
     then $|\bpsi|\! <\! |\bsigma|$ and $\text{diag}(\bpsi)\!\leq\!\text{diag}(\bsigma)$, and this last inequality is strict for at least one component along the diagonal.
    \item (Generalized variance matching) \\
      If $|\bpsi|\!=\! |\bsigma|$, then $\Psi_{ii}\! <\! \Sigma_{ii}$ and $\Psi^{-1}_{jj}\! <\! \Sigma^{-1}_{jj}$ for at least one $i$ and $j$.
  \end{enumerate}
\end{theorem} 
\end{mdframed}
The theorem shows, for instance, that a factorized approximation with the correct marginal variances will overestimate the generalized variance and underestimate at least one of the marginal precisions. It also highlights similar trade-offs when the factorized approximation matches the marginal precisions or the generalized variance. 


Our goal is to understand how different choices of divergences resolve the trade-offs imposed by \Cref{thm:impossibility}. To do so, we consider the case where $p$ and $q$ are Gaussian, as in eqs.~(\ref{eq:p_normal}--\ref{eq:q_normal}).
Note that in this setting $p$ and $q$ have the same generalized variance if and only if they have the same entropy.
All of the divergences in Table~\ref{tab:divergences} yield a correct estimate for the mean of $p$, but they yield different variational approximations for the diagonal covariance matrix $\bpsi$, and therefore different estimates of the marginal variances, marginal precisions, and entropy of $p$.
Moving forward, we say that one divergence \textit{dominates} another if it always returns a larger estimate of the marginal variances of $p$. 
%
\begin{mdframed}[hidealllines = true, backgroundcolor = Cerulean!10]
  \begin{definition}[Ordering of divergences]  \label{def:ordering}
     Let $\mathcal{P}$ and $\mathcal{Q}$ denote the families of multivariate Gaussian distributions in eqs.~(\ref{eq:p_normal}--\ref{eq:q_normal}), 
    where $\bpsi$ is diagonal but $\bsigma$ is not. 
    Consider two divergences $D_a$ and~$D_b$, and for any $p\!\in\!\mathcal{P}$, let $\bpsi(a)$ and $\bpsi(b)$ denote the covariances of the factorized approximations obtained, respectively, by minimizing $D_a(q,p)$ and $D_b(q,p)$ over all \mbox{$q\!\in\!\mathcal{Q}$}. We say $D_a$ dominates~$D_b$, written 
    $$D_a \succ D_b,$$
    if for any $p\in\mathcal{P}$ we have that $\text{diag}(\bpsi(a)) \ge \text{diag}(\bpsi(b))$ and also that
      $\Psi(a)_{ii}\!>\!\Psi(b)_{ii}$ for some element along the diagonal of these matrices.
\end{definition}
\end{mdframed}
%
The above definition\footnote{This definition can be further simplified if we impose the additional constraint that every row and column of $\bsigma$ has nonzero off-diagonal elements. In this case, we can adopt the simpler definition that $\mathcal{D}_a\succ\mathcal{D}_b$ if $\bpsi_a\succ\bpsi_b$, and all subsequent results hold with this more stringent characterization.} is based on an ordering of marginal variances, but it is also possible (because $\bpsi$ is diagonal) to define an equivalent ordering based on the marginal precisions. These orderings also imply an ordering of the generalized variances, though the converse is not true.

We now state the second main result of the paper, which is an ordering over the divergences listed in \Cref{tab:divergences}.
%
\begin{mdframed}[hidealllines = true, backgroundcolor = Cerulean!10]
  \begin{theorem}[Ordering theorem for FG-VI]  \label{thm:ordering}
  The divergences in Table~\ref{tab:divergences} are ordered in the sense given above. In particular, for any $(\alpha_1,\alpha_2)$ satisfying $0\!<\!\alpha_1\!<\!\alpha_2\!<\!1$, 
    \begin{equation}
     \SM(q||p) \prec \KL(q||p) \prec \text{D}_{\alpha_1}(p||q) \prec \text{D}_{\alpha_2}(p||q) \prec \KL(p||q) \prec \SM(p||q).
     \label{eq:ordering}
   \end{equation}
  \end{theorem}
\end{mdframed}
 The theorem shows, for instance, that the (reverse) score-based divergence yields smaller estimates of marginal variance than the (reverse) KL divergence, which in turn yields smaller estimates than all of the $\alpha$-divergences. We prove \Cref{thm:ordering} in \Cref{sec:ordering}. The main challenge here arises from the fact that the variational approximations from the score-based and $\alpha$-divergences cannot be computed in closed form, and thus it is necessary (as done in \Cref{sec:divergence}) to characterize these solutions in other ways that can be further analyzed.


The proof of \Cref{thm:ordering} also yields several intermediate results of interest. Most notably, we obtain the following positive results for VI-based uncertainty quantification:
\begin{itemize}[itemsep=-1ex]
\item[(i)] $q$ matches the marginal variances of $p$ when minimizing $\KL(p||q)$ \citep{Mackay:2003}; 
\item[(ii)] $q$ matches the marginal precisions of $p$ when minimizing $\KL(q||p)$ \citep{Turner:2011}; 
\item[(iii)] $q$ matches the generalized variance (or entropy) of~$p$ when minimizing $\Ra(p||q)$, for some $\alpha \in (0, 1)$, which depends on $\bsigma$.
\end{itemize}
We also obtain certain cautionary results. For example, we show that~$q$ can wildly misestimate uncertainty when minimizing a score-based divergence; in fact, the estimated variance may collapse to 0 when optimizing $S(q||p)$ or blow up to infinity when optimizing $S(p||q)$. We refer to such cases as instances of \textit{variational collapse}, where the variational approximation is not well-defined because \mbox{$\arg\!\inf_{q\in\mathcal{Q}} D(q,p)\not\in\mathcal{Q}$}.

We extend some of our results for $\alpha$-divergences to the case where $\alpha\! >\! 1$. 
In this regime, we prove the following:  (i) $\Ra(p||q) \succ \KL(p||q)$, (ii) there does \textit{not} exist an ordering of $\Ra(p||q)$ and $S(p||q)$, and (iii) $\Ra(p||q)$ yields strictly positive, bounded estimates of the variance and 
so (unlike the score-based divergences) it does not suffer from variational collapse.
We are \textit{not} able to show for $\alpha\!>\!1$ that the $\alpha$-divergences are ordered amongst themselves, satisfying $\text{D}_{\alpha_2}(p||q) \!\succ\! \text{D}_{\alpha_1}(p||q)$ when $\alpha_2\! >\! \alpha_1\! >\! 1$.
As we discuss later, there are technical reasons why our proof for $\alpha\!\in\!(0,1)$ does not extend to this case where $\alpha\!>\!1$. Based on our numerical experiments, however, we conjecture that the $\alpha$-divergences are indeed ordered amongst themselves for $\alpha\! >\! 1$, and we leave the proof (or disproof) of this conjecture for future work.

For convenience, Table~\ref{tab:notations} provides a summary of the notation used in the paper.

\begin{table}[]
    \centering
    \def\arraystretch{1.5}
    \begin{tabular}{r|l}
         \rowcolor{Cerulean!10} {\bf Notation} & {\bf Definition}  \\
         F-VI & Factorized Variational Inference \\
         \rowcolor{Cerulean!10} FG-VI & Factorized Gaussian Variational Inference \\
         diag($\bf A$) & The vector formed from the diagonal of the squared matrix $\bf A$. \\
         \rowcolor{Cerulean!10} ${\bf v} \ge {\bf u}$ & For two vectors ${\bf u}, {\bf v} \in \mathbb R^n$, $v_i \ge u_i, \forall i$. \\
         $A^{-1}_{ij}$ & The $(ij)^\text{th}$ element of ${\bf A}^{-1}$.
         In general $A^{-1}_{ij} \neq 1/A_{ij}$. \\
         \rowcolor{Cerulean!10} ${\bf A}\succ{\bf B}$ & 
         The matrix ${\bf A}\!-\!{\bf B}$ is positive definite. \\
         $||{\bf v}||^2_{\bf A}$ & For ${\bf v} \in \mathbb R^n$ and ${\bf A} \in \mathbb R^{n \times n}$, the weighted inner-product ${\bf v}^T {\bf A} {\bf v}$. \\
         \rowcolor{Cerulean!10} $D_a \succ D_b$ & Divergence $D_a$ \textit{dominates} divergence $D_b$ (Definition~\ref{def:ordering}). \\
         $\bsigma$ & The covariance matrix of the target $p$ (eq.~\ref{eq:p_normal}). \\
         \rowcolor{Cerulean!10} $\bpsi$ & The covariance matrix of the approximation $q$ (eq.~\ref{eq:q_normal}). \\
         
    \end{tabular}
    \caption{{Summary of notation. For the notation on divergences, see Table~\ref{tab:divergences}.}}
    \label{tab:notations}
\end{table}

\subsection{Related Work}

Our work is in line with several efforts to evaluate  VI through the lens of downstream inferential tasks.
The mean and variances of $p$ are of most interest for Bayesian statistics, but
\citet{Huggins:2020} caution that a low $\KL(q||p)$ divergence does not guarantee that $q$ accurately estimates these quantities.
Instead they recommend to estimate the Wasserstein distance, which can be used to bound the difference in moments of $q$ and~$p$; see also \citet{Biswas:2024}.
In latent variable models, VI is used to bound a marginal likelihood which is then maximized with respect to the model parameters, e.g the weights of a neural network \citep{Tomczak:2022}.
Several works examine the tightness of this bound \citep[e.g][]{Li:2016, Dieng:2017, Daudel:2023b} and the statistical properties of the resulting maximum likelihood estimator \citep{Wang:2018}.
VI is also used in tandem with other inference schemes---for example, to initialize Markov chain Monte Carlo \citep{Zhang:2022} or to estimate a proposal distribution for importance sampling. 
For the latter, the quality of VI can be assessed from the distribution of importance weights \citep{Yao:2018, Vehtari:2024}.
In sum, there are many different ways to evaluate VI, and one lesson of our paper is that different inferential goals may be incompatible---may, in fact, compete against one another.

Many researchers have studied, both theoretically and empirically, how well F-VI estimates uncertainty (and, in particular, the marginal variances) when minimizing $\KL(q||p)$ \citep[e.g.,][]{Mackay:2003, Wang:2005, Bishop:2006, Turner:2011, Giordano:2018}.
Several works also examine the more specific setting in \Cref{sec:divergence} where $p$ and $q$ are Gaussian with (respectively) a non-diagonal and diagonal covariance matrix.
\citet{Mackay:2003} reports that $q$ matches the marginal variances when minimizing $\KL(p||q)$, while \citet{Turner:2011} find that $q$ matches the marginal precisions when minimizing $\KL(q||p)$.
In a precursor to this paper, \citet{Margossian:2023} prove that a precision-matching solution underestimates both the entropy and the marginal variances; here we provide a new and more intuitive proof of this result. We also note that  \Cref{thm:impossibility} provides a more general framework for understanding these sorts of trade-offs, as it assumes neither that $p$ and $q$ are Gaussian nor that $q$ is obtained by minimizing a particular divergence, such as $\KL(q||p)$.

Previous studies in VI have investigated a large class of divergences for the optimization in eq.~(\ref{eq:argminDqp}). A number of authors have considered the $\alpha$-divergence~\citep{Li:2016, Dieng:2017, Daudel:2021}, for instance, as it provides a non-trivial generalization of the KL divergence. More recently, \citet{Cai:2024} introduced the score-based divergence $S(q||p)$ in Table~\ref{tab:divergences} and showed that it can be efficiently optimized for Gaussian variational families; see also \citet{Modi:2025}.
It is known that the $\alpha$-divergence reduces in certain limits to the KL divergence (either from $q$ to $p$, or vice versa); however, there is no such correspondence for the score-based divergences, and thus it is not obvious how the variational approximations obtained by minimizing $S(q||p)$ or $S(p||q)$ might compare to those obtained by minimizing $\KL(q||p)$ or $\KL(p||q)$. Our analysis, notably the ordering in \Cref{thm:ordering}, sheds light on the matter. 

Finally, we highlight a recent contribution by \citet{Chen:2025} who build on the ideas in this paper to analyze the weighted Fisher divergence, defined as
\begin{equation}
    S_M(q||p) = \mathbb E_q \left [ \nabla \left \|\log \frac{q(\bz)}{p(\bz)} \right \|^2_{\bf M} \right]
\end{equation}
for ${\bf M} \succ 0$.
If ${\bf M} = \bpsi$, then this divergence reduces to the reverse score-based divergence $S(q||p)$; on the other hand, if ${\bf M} = {\bf I}$ (the identity matrix), then $S_M(q||p)$ reduces to the Fisher divergence, which we denote by $F(q||p)$.
\citet{Chen:2025} notably show that (i) $F(q||p) \prec \KL(q||p)$, (ii)  $F(q||p)$ and $S(q||p)$ cannot be ordered, and (iii) unlike the score-based divergence $S(q||p)$, the Fisher divergence $F(q||p$) is not subject to variational collapse.

\section{Proof of Impossibility Theorem}
\label{sec:impossibility}

We begin by restating the impossibility theorem of the previous section as a collection of trade-offs.

\begin{mdframed}[hidealllines = true, backgroundcolor = Cerulean!10]
\begin{theorem}[Restatement of impossibility theorem]
Let $p$ and $q$ be distributions over~$\mathbb{R}^N$ with covariances $\bsigma$ and $\bpsi$, respectively, where $\bpsi$ is diagonal but $\bsigma$ is not.
\begin{enumerate}  
    \item \textbf{Trade-off between marginal variances and generalized variance:} \\
      If $\text{diag}({\bpsi})\!=\!\text{diag}({\bsigma})$, then $|\bpsi|\! > \!|\bsigma|$, and 
      if $|\bpsi|\!=\!|\bsigma|$, then $\Psi_{ii}\!<\!\Sigma_{ii}$ for some $i$.
    \item \textbf{Trade-off between marginal precisions and generalized variance:} \\
      If $\text{diag}({\bpsi^{-1}})\!=\!\text{diag}({\bsigma^{-1}})$, then $|\bpsi|\! < \!|\bsigma|$, and 
      if $|\bpsi|\!=\!|\bsigma|$, then $\Psi^{-1}_{ii}\!<\!\Sigma^{-1}_{ii}$\! for some $i$.
    \item \textbf{Variance-precision trade-off:} \\
    If $\text{diag}(\bpsi)\!=\!\text{diag}(\bsigma)$, then $\text{diag}(\bpsi^{-1})\!\leq\!\text{diag}(\bsigma^{-1})$
    and $\Psi^{-1}_{ii}\!<\!\Sigma^{-1}_{ii}$ for some $i$. \\
    If $\text{diag}(\bpsi^{-1})\!=\!\text{diag}(\bsigma^{-1})$,~then $\text{diag}(\bpsi)\!\leq\!\text{diag}(\bsigma)$ and $\Psi_{ii}\!<\!\Sigma_{ii}$ for some $i$. 
\end{enumerate}
\label{thm:restate}
\end{theorem}
\end{mdframed}

\noindent
The above restatement makes exactly the same claims as \Cref{thm:impossibility}, but the proof is easier to organize along these lines.

We emphasize that the trade-offs listed above are not tied to any particular choice of the divergence $D(q,p)$ in eq.~(\ref{eq:argminDqp}).
Rather, these trade-offs are inherent to the use of a factorized approximation, and they arise regardless of how $q$ is chosen. Indeed, we will prove each trade-off by appealing only to the properties of $\bsigma$ and $\bpsi$ as positive-definite matrices. However, for each trade-off, we will also provide additional intuitions (and in one case, an additional proof) that reflect the statistical setting in which these matrices arise.


\vspace{2ex}
\noindent
\fcolorbox{Cerulean!10}{Cerulean!10}{\textit{Proof of trade-off between marginal variances and generalized variance.}} 
We begin by defining two auxiliary matrices 
whose elements are dimensionless (unlike those of $\bsigma$ and $\bpsi$).  The first of these is the \textit{correlation} matrix $\bC\in \mathbb R^{n \times n}$ of the target distribution~$p$, with elements
  \begin{equation} \label{eq:correlation}
    C_{ij} = \frac{\Sigma_{ij}}{\sqrt{\Sigma_{ii} \Sigma_{jj}}},
  \end{equation}
  which is positive-definite
  and has unit diagonal entries, $C_{ii}\!=\!1$. The second is the diagonal matrix $\bR\in\mathbb{R}^{n\times n}$ that records the \textit{ratios} of marginal variances in $q$ and $p$; i.e., 
   \begin{equation}
   \label{eq:ratio}
    R_{ii} = \frac{\Psi_{ii}}{\Sigma_{ii}}.
  \end{equation}
  In what follows, it will be convenient to consider the generalized variances on a log scale. Taking differences, we have 
\begin{equation}
\log |\bsigma| - \log |\bpsi|
 = \log\left|\bpsi^{-\frac{1}{2}}\bsigma\bpsi^{-\frac{1}{2}}\right|
 = \log\left|\bR^{-\frac{1}{2}}\bC\bR^{-\frac{1}{2}}\right|
 = \log|\bR|^{-1} - \tfrac{1}{2}\log|\bC|^{-1}.
 \label{eq:RC}
\end{equation}
The first determinant on the right side of eq.~(\ref{eq:RC}) is easy to compute because the matrix $\bR$, as defined in eq.~(\ref{eq:ratio}), is diagonal. We can bound the second determinant by the Hadamard inequality~\citep[Theorem 7.8.1]{Horn:2012}, which states that 
\begin{equation}
|\bC|<\prod_{i=1}^n C_{ii} = 1.
\label{eq:HadamardC}
\end{equation}
The inequality in eq.~(\ref{eq:HadamardC}) is strict under the assumption that $\bsigma$ (and hence also~$\bC$) is not diagonal. Combining the last two results, we see that
\begin{equation}
    \log|\bsigma| - \log |\bpsi| \ <\ 
         \sum_i\log\left(\frac{\Sigma_{ii}}{\Psi_{ii}}\right).
\label{eq:diffH}
\end{equation}
Now suppose that $\text{diag}(\bpsi)\!=\!\text{diag}(\bsigma)$; then from eq.~(\ref{eq:diffH}) we see that $|\bsigma|\!<\!|\bpsi|$, proving the first part of the trade-off. Conversely, suppose that $|\bsigma|\!=\!|\bpsi|$; then from eq.~(\ref{eq:diffH}) we see that $\sum_i \log\Sigma_{ii}\!>\!\sum_i \log\Psi_{ii}$, and this can only be true if $\Sigma_{ii}\!>\!\Psi_{ii}$ for some $i$. This proves the second part of the trade-off. \mbox{\hspace{65ex}$\blacksquare$}

\vspace{2ex}
The above trade-off 
has a particularly interesting interpretation in the case where $p$ and~$q$ are Gaussian, as in eqs.~(\ref{eq:p_normal}--\ref{eq:q_normal}). In this case, the generalized variance of $q$, given by $|\bpsi|$, provides effectively the same measure of uncertainty as the \textit{entropy} of $q$, which is equal to $\frac{1}{2}\log|\bpsi|$ plus an additive constant. Thus when $q$ correctly estimates the marginal variances, it overestimates the entropy, and when it correctly estimates the entropy, it underestimates at least one marginal variance. The entropy of $q$ in this setting is governed by a \textit{shrinkage-delinkage trade-off} \citep{Margossian:2023}: the entropy of $q$ is decreased when $q$ underestimates the marginal variances of $p$ (the shrinkage effect), but it is increased when $q$ ignores the correlations in $p$ between off-diagonal components (the delinkage effect).


\vspace{2ex}
\noindent
\fcolorbox{Cerulean!10}{Cerulean!10}{\textit{Proof of trade-off between marginal precisions and generalized variance.}}
This \\ proof follows the same structure as the preceding one. We define auxiliary matrices $\tilde{\bC}$ and~$\tilde{\bR}$ with the dimensionless elements
\begin{equation}
   \tilde C_{ij} = \frac{\Sigma^{-1}_{ij}}{\sqrt{\Sigma^{-1}_{ii} \Sigma^{-1}_{jj}}}, 
   \quad\quad
   \tilde R_{ii} = \frac{\Psi_{ii}^{-1}}{\Sigma^{-1}_{ii}},
   \label{eq:tildeCR}
\end{equation}
where $\tilde{\bC}$ has all ones along the diagonal and $\tilde{\bR}$ has all zeroes off the diagonal. In terms of these matrices, in analogy to eq.~(\ref{eq:RC}), we can write
\begin{equation}
\log |\bsigma| - \log |\bpsi| = \log\left|\bR^{-\frac{1}{2}}\bC\bR^{-\frac{1}{2}}\right| = \log\left|\tilde{\bR}^{\frac{1}{2}}\tilde{\bC}^{-1}\tilde{\bR}^{\frac{1}{2}}\right| = \log|\tilde{\bR}| - \tfrac{1}{2}\log|\tilde{\bC}|.
\label{eq:Htilde}
\end{equation}
As before, the determinant of $\tilde{\bR}$ is easy to compute because it is a diagonal matrix, and again we deduce from the Hadamard inequality that $|\tilde{\bC}|<\prod_{i} \tilde{\bC}_{ii} = 1$. Thus we find
\begin{equation}
\log |\bsigma|- \log |\bpsi|\ >\ \sum_i \log\left(\frac{\Psi_{ii}^{-1}}{\Sigma^{-1}_{ii}}\right).
\label{eq:diffH_lb}
\end{equation}
Note that the direction of inequality in eq.~(\ref{eq:diffH_lb}) is reversed from
that in eq.~(\ref{eq:diffH}). Now suppose that $\text{diag}(\bpsi^{-1})\!=\!\text{diag}(\bsigma^{-1})$; then from eq.~(\ref{eq:diffH_lb}) we see that $|\bsigma|\!>\!|\bpsi|$, proving the first part of the trade-off. Conversely, if $|\bsigma|\!=\!|\bpsi|$, then from eq.~(\ref{eq:diffH_lb}) we see that $\sum_i \log\Sigma^{-1}_{ii}\!>\!\sum_i \log\Psi^{-1}_{ii}$, and this can only be true if $\Sigma^{-1}_{ii}\!>\!\Psi^{-1}_{ii}$ for some $i$. This proves the second part of the trade-off.~\mbox{\hspace{57ex}$\blacksquare$}

\vspace{2ex}
We can also interpret this trade-off in the special case when $p$ and $q$ are Gaussian, as in eqs.~(\ref{eq:p_normal}--\ref{eq:q_normal}). In this case, it states the following: when $q$ correctly estimates the marginal precisions, it underestimates the entropy, and when it correctly estimates the entropy, it underestimates at least one marginal precision.

   

Next we prove the final trade-off between marginal variances and precisions. A short proof of this trade-off was given in \citet{Margossian:2023}. Here instead we appeal to a stronger result from linear algebra~\citep[Theorem 7.17]{Horn:2012}, which will also prove 
useful in \Cref{sec:ordering}. This result is given by the following lemma.

  \vspace{1ex}
  \begin{mdframed}[hidealllines = true, backgroundcolor = Cerulean!10]
  \begin{lemma}
  \label{prop:hadamard}
  Let $\mathbf{A}\in\mathbb{R}^{n\times n}$. If $\mathbf{A}\!\succ\!\mathbf{0}$, then
  \vspace{-2ex}
  \begin{align}
      A_{ii}A^{-1}_{ii} &\geq 1\quad\mbox{for all $i$}, \label{eq:Ageq}\\
      A_{jj}A^{-1}_{jj} &> 1\quad\mbox{whenever $\sum_{k\neq j}|A_{jk}|>0$}.
      \label{eq:Agt}
  \end{align}
 \end{lemma}
  \end{mdframed}
  \vspace{1ex}
Any non-diagonal element $A_{jk}$ in this lemma generates strict inequalities of the form in eq.~(\ref{eq:Agt}) for the diagonal elements $A_{jj}$ and $A_{kk}$.
From this lemma, we obtain the variance-precision trade-off
by substituting either the covariance matrix $\bsigma$ or the precision matrix $\bsigma^{-1}$ in Theorem~\ref{thm:restate} for the matrix $\mathbf{A}$ in
eqs.~(\ref{eq:Ageq}--\ref{eq:Agt}). 
Unfortunately, neither the above lemma 
nor the proof in \citet{Margossian:2023} offer much intuition.
Here, we propose an alternative proof rooted in statistical notions.


\vspace{2ex}
\noindent
\fcolorbox{Cerulean!10}{Cerulean!10}{\textit{Proof of variance-precision trade-off.}}
Let $\xi\sim\mathcal{N}(0,\bsigma)$, and let $\bxi_{-i}$ denote all the elements of $\bxi$ other than $\xi_i$. 
Also denote the density of $\xi$ by $\pi$. From the conditional distribution
\begin{equation}
    \pi(\xi_i \mid \bxi_{-i}) \propto \exp\left(-\frac{1}{2}\xi_i\Sigma^{-1}_{ii}\xi_i - \xi_i\sum_{j\neq i}\Sigma^{-1}_{ij}\xi_j\right),
\end{equation}
we can identify (from the first quadratic term in the exponent) that $\text{Var}[\xi_i|\bxi_{-i}] = 1/\Sigma^{-1}_{ii}$; in other words, the conditional variance of this variable is given by the reciprocal of its marginal precision. Now by the law of total variance, we have that
\begin{align}
\Sigma_{ii}
  &= \text{Var}[\xi_i], \\
  &= \mathbb{E}[\text{Var}[\xi_i|\bxi_{-i}]] + \text{Var}[\mathbb{E}[\xi_i|\bxi_{-i}]], \\
  &\geq \mathbb{E}[\text{Var}[\xi_i|\bxi_{-i}]], \\
  &= \mathbb{E}[1/\Sigma^{-1}_{ii}], \\
  &= 1/\Sigma^{-1}_{ii}.
\end{align}

%
\noindent
Thus we have shown $\Sigma_{ii}\Sigma^{-1}_{ii}\geq 1$ for all $i$. Moreover, since by assumption $\bsigma$ is not diagonal, there must exist some coordinate such that $\mathbb E[\xi_i|\bxi_{-i}]\neq\mathbb{E}[\xi_i]$ and $\text{Var}[\mathbb E[\xi_i|\bxi_{-i}]] > 0$; for this coordinate, the inequality is strict with $\Sigma_{ii}\Sigma^{-1}_{ii}> 1$.
Suppose now that $q$ matches the marginal precisions of $p$; then for all elements along the diagonal, we have
\begin{equation}
    \Psi_{ii} = (\Psi^{-1}_{ii})^{-1} = (\Sigma^{-1}_{ii})^{-1} \leq \Sigma_{ii},
\end{equation}
and this inequality must be strict for at least one of the marginal variances. This proves one part of the trade-off. Likewise, if $q$ matches the marginal variances of $p$, then for all elements along the diagonal we have
\begin{equation}
    \Psi^{-1}_{ii} = (\Psi_{ii})^{-1} = (\Sigma_{ii})^{-1} \leq \Sigma^{-1}_{ii},
\end{equation}
and this inequality must be strict for at least one of the marginal precisions. This proves the other part of the trade-off.~\mbox{\hspace{63ex}$\blacksquare$}

\section{Divergences for FG-VI}  \label{sec:divergence}

\begin{table}
  \begin{center}
    \renewcommand{\arraystretch}{1.5}
 \addtolength{\tabcolsep}{1ex}
     \begin{tabular}{rll} 
     \rowcolor{Cerulean!10} & {\bf Reduced divergence} & {\bf (Implicit) Solution}  \\
    $\KL(q || p)$ & $\frac{1}{2} \text{trace} \left ( \bpsi \bsigma^{-1} \right) - \log |\bpsi \bsigma^{-1}| + c$ & $\Psi_{ii} = 1 / \Sigma^{-1}_{ii}$ \\
      \rowcolor{Cerulean!10} 
    $\KL(p || q)$ & $\frac{1}{2} \sum_{i = 1}^n (\log \frac{\Psi_{ii}}{\Sigma_{ii}} + \frac{\Sigma_{ii}}{ \Psi_{ii}}) + c$ & $\Psi_{ii} = \Sigma_{ii}$ \\
    $\Ra (p || q)$ & $\frac{1}{\alpha (\alpha - 1)} \left [ |\bsigma_\alpha|^{-\frac{1}{2}} |\bpsi|^\frac{\alpha}{2} |\bsigma|^\frac{1 - \alpha}{2} - 1 \right]$ & $\Psi_{ii} = \left [ \alpha \bsigma^{-1} + (1 - \alpha) \bpsi^{-1} \right]^{-1}_{ii}$ \\
    \rowcolor{Cerulean!10} 
    $\SM (q || p)$ & $\text{trace}\left({\bf I}\! -\! \bpsi \bsigma^{-1}\right)$ & $\underset{\bs \ge 0}{\text{argmin}} \left [ \frac{1}{2} \bs^T \bH \bs - {\bf 1}^T \bs \right]$, $s_{ii} = \Psi_{ii} \Sigma^{-1}_{ii}$ \\
    $\SM (p || q)$ & $\text{trace} \left({\bf I}\! -\! \bsigma \bpsi^{-1}\right) $  & $\underset{\bt \ge 0}{\text{argmin}} \left [ \frac{1}{2} \bt^T {\bf J} \bt - {\bf 1}^T \bt \right]$, $t_{ii} = \Psi^{-1}_{ii} \Sigma_{ii}$
    \end{tabular}
 \addtolength{\tabcolsep}{-1ex}  
    \caption{{Divergences when $p$ and $q$ are Gaussian (eq~\ref{eq:p_normal}-\ref{eq:q_normal}). 
    We report the reduced divergence, obtained after matching the means of $p$ and $q$.
    For the R\'enyi divergence, the optimal covariance can be found by solving a fixed-point equation.
    For the score-based divergences, we need to solve a non-negative quadratic program.
    The auxiliary matrix $\bsigma_\alpha$ is defined in \cref{eq:sigma_alpha}, $\bH$ in \cref{eq:H}, and ${\bf J}$ in \cref{eq:J}.
    }}
    \label{tab:FG-VI}
  \end{center}
\end{table}

In this section we derive the solutions for FG-VI that are obtained from eq.~(\ref{eq:argminDqp}--\ref{eq:q_normal}) by minimizing the divergences in Table~\ref{tab:divergences}.  We also highlight the properties of solutions that are obtained in this way. As a useful reference, we summarize the results for these solutions in Table~\ref{tab:FG-VI}. We emphasize that all of the results in this section are based on the further assumption that $p$ and $q$ are Gaussian, as in eqs.~(\ref{eq:p_normal}--\ref{eq:q_normal}).


\subsection{KL Divergences}
\label{sec:KL}

The KL divergence is not a symmetric function of its arguments, and thus the solutions for FG-VI depend on the direction of the divergence that is minimized. We consider each direction in turn.

\vspace{2ex}
\noindent
\fcolorbox{Cerulean!10}{Cerulean!10}{\textit{Reverse direction: $\KL(q||p)$.}} Most VI is based on minimizing the reverse KL divergence in Table~\ref{tab:divergences}. The implications of this choice have been extensively studied for FG-VI. For completeness, we review certain key results in the notation of this paper. 
When $q$ is chosen to minimize $\KL(q||p)$, it matches the mean of $p$ (that is, $\bnu\!=\!\bmu$), and its 
covariance is given~by
\begin{equation}
 \textrm{diag}(\bpsi^{-1}) = \textrm{diag}(\bsigma^{-1}).
 \label{eq:precision-matching}
\end{equation}
Further details of these calculations can be found in the references~\citep{Turner:2011, Margossian:2023}. 
We summarize the main properties of this solution for uncertainty quantification, all of which follow from the results of the previous section:
\begin{itemize}[itemsep=-0.5ex]
   \item[(i)] The marginal precisions of $q$ match those of $p$.
     \item[(ii)] The entropy of $q$ underestimates the entropy of $p$.
        \item[(iii)] The marginal variances of $q$ underestimate the marginal variances of $p$. 
   \item[(iv)] The marginal variances of $q$ match the conditional variances of $p$ when each Gaussian random variable is conditioned on all the others.
\end{itemize}

It has been widely observed that VI based on the reverse KL divergence tends to underestimate uncertainty~\citep{Mackay:2003, Minka:2005, Turner:2011, Blei:2017, Giordano:2018}. 
The above results show that the ``uncertainty deficit'' of FG-VI depends on the measure that is used to quantify uncertainty. Indeed, when $p$ is Gaussian, the marginal precisions are correctly estimated by minimizing $\KL(q||p)$; moreover, it has been shown that when $\bsigma$ has \textit{constant} off-diagonal entries, the entropy gap between $p$ and $q$ only grows as $\mathcal O(\log n)$, while the entropy itself grows as $\mathcal O(n)$, meaning that the fractional entropy gap tends to 0 as $n\! \to\! \infty$ \citep[Theorem 3.6]{Margossian:2023}.
Crucially, for~$q$ to estimate the entropy well, it must necessarily underestimate the marginal variances, as prescribed by \cref{eq:RC}. 
Other asymptotically correct results are obtained in the thermodynamic limit of Ising models with constant, long-range interactions, where the mean-field approximation has its roots \citep{Parisi:1988}.
  

\vspace{2ex}
\noindent
\fcolorbox{Cerulean!10}{Cerulean!10}{\textit{Forward direction: $\KL(p||q)$.}} Next we consider the forward KL divergence in Table~\ref{tab:divergences}. This divergence is not generally minimized for VI because it involves an expectation, namely $\mathbb{E}_p[\log(p/q)]$ with respect to the target distribution~$p$. When $p$ is not tractable, it is not possible to compute this expectation analytically, and it may be too expensive to estimate this expectation by sampling. 
%
Still, it is illuminating to contrast the properties of the reverse and forward KL divergences for FG-VI. 
The latter is similarly minimized by setting $\bnu\!=\!\bmu$, but now we obtain a different solution when the remaining terms are minimized with respect to~$\bpsi$. In particular, $\KL(p||q)$ is minimized by setting 
\begin{equation}
    \textrm{diag}(\bpsi) = \textrm{diag}(\bsigma),
    \label{eq:variance-matching}
\end{equation}
thus matching the marginal distributions of $p$ and $q$ \citep{Mackay:2003}.
We summarize the main properties of this solution for uncertainty quantification, all of which follow directly from the Impossibility Theorem:
\begin{itemize}[itemsep=-0.5ex]
   \item[(i)] The marginal variances of $q$ match those of $p$. 
     \item[(ii)] The entropy of $q$ overestimates the entropy of $p$.
        \item[(iii)] The marginal precisions of $q$ underestimate the marginal precisions of $p$.
\end{itemize}
Comparing the results for $\KL(p||q)$ and $\KL(q||p)$, we see the important effects of the divergence when VI is used for uncertainty quantification. These effects motivate our subsequent study of additional divergences.

\subsection{$\alpha$-divergence}

Next we consider a one-parameter family of divergences that includes the KL divergences in the previous section as limiting cases. The $\alpha$-divergence is given by
 \begin{equation} \label{eq:alpha-divergence}
      \Ra (p || q) = \frac{1}{\alpha(\alpha - 1)} \int \ \left (\frac{p^\alpha(\bz)}{q^\alpha(\bz)} - 1 \right) \, q(\bz)\,
 \text d \bz,
\end{equation}
where it is assumed that $\alpha\!>\!0$ and $\alpha\!\neq\!1$. Like the KL divergence, the $\alpha$-divergence has the property that $\Ra(p||q) \ge 0$, with equality holding if and only $p\!=\!q$. 

The $\alpha$-divergence in eq.~(\ref{eq:alpha-divergence}) has been studied in the context of VI~\citep{Li:2016}.
The divergence can be defined in various ways, but these various definitions\footnote{Closely related is the R\'enyi divergence, given by $R_\alpha(p||q) = (\alpha\!-\!1)^{-1}\log\int p(\bz)^\alpha q(\bz)^{1-\alpha}$.} have in common that they all yield the same underlying optimization for VI. We use the above definition \citep{Cichocki:2010} to recover the KL divergences in the previous sections as limiting cases:
    \begin{align} 
      \label{eq:alpha-limit0}
      \lim_{\alpha \to 0} \Ra(p || q) & = \KL(q || p), \\
      \label{eq:alpha-limit1}
      \lim_{\alpha \to 1} \Ra(p || q) & = \KL(p || q).
    \end{align}
Thus for $\alpha\!\in\!(0,1)$, the $\alpha$-divergences provide a one-parameter family of divergences interpolating between $\KL(q||p)$ and $\KL(p||q)$. Likewise, for $\alpha\!=\!2$, eq.~(\ref{eq:alpha-divergence}) recovers the $\chi^2$-divergence, which can also be minimized for approximate inference \citep{Dieng:2017}.

The $\alpha$-divergence can be computed exactly between two multivariate Gaussians~\citep{Burbea:1984,Liese:1987,Hobza:2009,Gil:2013} such as the ones in \mbox{eqs.~(\ref{eq:p_normal}--\ref{eq:q_normal})}.  To do so, it is convenient to define the matrices
\begin{align}
\bsigma_\alpha &= \alpha\bpsi + (1\!-\!\alpha)\bsigma, \label{eq:sigma_alpha}\\
\bphi_\alpha &= \alpha\bsigma^{-1} + (1\!-\!\alpha)\bpsi^{-1}, \label{eq:phi_alpha}
\end{align}
which are also related by the identity $\bsigma_\alpha = \bsigma\,\bphi_\alpha\bpsi$. It is known that the integral for $\Ra(p||q)$ in eq.~(\ref{eq:alpha-divergence}) only exists when
$\bphi_\alpha \succ 0$. In this case, the $\alpha$-divergence is given by
   \begin{align}
    \label{eq:renyi-gaussian}
    \Ra(p||q) &= \frac{1}{\alpha(\alpha\!-\!1)}\left[
      e^{-\frac{\alpha(1-\alpha)}{2}(\bmu-\bnu)^\top \bsigma_\alpha^{-1}(\bmu-\bnu)}\, 
         |\bsigma_\alpha|^{-\frac{1}{2}}|\bpsi|^{\frac{\alpha}{2}}|\bsigma|^{\frac{1-\alpha}{2}} - 1\right].
\end{align}
This expression vanishes if $\bmu\!=\!\bnu$ and $\bpsi\!=\!\bsigma$ (in which case $\bsigma_\alpha\!=\!\bsigma$), but not otherwise.

To analyze FG-VI with this divergence, we must minimize eq.~(\ref{eq:renyi-gaussian}) with respect to the variational mean $\bnu$ and diagonal covariance $\bpsi$ of $q$. The next propositions establish useful properties of the variational approximations that are found in this way. In particular, the first shows that $q$ matches the mean of $p$.


\begin{mdframed}[hidealllines=true, backgroundcolor = Cerulean!10]
\begin{proposition}[Mean matching]
Let $\alpha \in \mathbb R^+ \backslash \{0, 1\}$, and let $p$ and~$q$ be given by eqs.~(\ref{eq:p_normal}--\ref{eq:q_normal}) where $\bpsi$ is diagonal.
If $q$ minimizes the $\alpha$-divergence in eq.~(\ref{eq:alpha-divergence}), then it matches the mean of $p$; that is, $\bnu\!=\! \bmu$.
\end{proposition}
\end{mdframed}

\begin{proof}
  We proceed by optimizing the right side of eq.~(\ref{eq:renyi-gaussian}).
  First, we consider the case where $\alpha\!\in\!(0,1)$.
  In this case, the prefactor $1 / \alpha(\alpha\!-\!1)$ is \textit{negative}, and the expression as a whole is minimized by \textit{maximizing} the first term in brackets. Upon taking logarithms, we see equivalently that
\begin{equation} \label{eq:alpha-objective}
\argmin_{q\in\mathcal{Q}}\Big[\Ra(p||q)\Big] = \argmin_{\bnu,\bpsi}\left[\alpha(1\!-\!\alpha)(\bmu\!-\!\bnu)^\top \bsigma_\alpha^{-1}(\bmu\!-\!\bnu) + \log|\bsigma_\alpha| - \alpha\log|\bpsi|\right].
\end{equation}
The first term on the right is minimized (and zeroed) by setting $\bnu\!=\!\bmu$, thus matching the means of $q$ and $p$.

A similar proof holds when $\alpha > 1$.
\end{proof}

The next proposition shows that $q$ estimates finite, non-zero variances, or equivalently, that $\bpsi\!\succ\! 0$ and $\bpsi^{-1}\!\succ\! 0$ for any $\bpsi$ minimizing the $\alpha$-divergence in eq.~(\ref{eq:renyi-gaussian}).


\begin{mdframed}[hidealllines=true, backgroundcolor = Cerulean!10]
\begin{proposition}[Variance bounds]
\label{prop:variance-bounds}
   Let $\alpha \in \mathbb R^+ \backslash \{0, 1\}$, and let $p$ and~$q$ be given by eqs.~(\ref{eq:p_normal}--\ref{eq:q_normal}) where $\bpsi$ is diagonal. If $q$ minimizes the $\alpha$-divergence in eq.~(\ref{eq:alpha-divergence}), then its variances are strictly positive and finite; that is, $0\! <\! \Psi_{ii}\! <\! \infty$ for all $i$.
\end{proposition}
\end{mdframed}
\begin{proof}
Again we proceed by minimizing the right side of eq.~(\ref{eq:renyi-gaussian}). We first consider the case $\alpha\! \in\! (0, 1)$.
By the previous proposition, the minima occurs at $\bnu\! =\! \bmu$. Making this substitution, we find
\begin{align} 
    \argmin_{q \in \mathcal Q} \left [ \Ra(p||q) \right] & = \argmin_{\bpsi} \frac{1}{\alpha (\alpha\! -\! 1)}\, |\bsigma_\alpha|^{-\frac{1}{2}} |\bpsi|^\frac{\alpha}{2} \\
    & = \argmax_{\bpsi} |\bsigma_\alpha|^{-\frac{1}{2}}\,|\bpsi|^\frac{\alpha}{2} \\
    & = \argmax_{\bpsi}\big|\alpha \bpsi + (1\!-\!\alpha) \bsigma \big|^{-\frac{1}{2}}\, \big|\bpsi^{-\alpha}\big|^{-\frac{1}{2}} \\
    & = \argmin_{\bpsi} \Big|\alpha \bpsi^{1 - \alpha} + (1\! -\! \alpha) \bsigma \bpsi^{-\alpha}\Big|. \label{eq:argmin-alpha}
\end{align}
The determinant in eq.~(\ref{eq:argmin-alpha}) is finite for any diagonal covariance $\bpsi$ satisfying $\Psi\!\succ\!0$ and $\Psi^{-1}\!\succ\! 0$. But for $\alpha\!\in(0,1)$, this determinant diverges if any $\Psi_{ii}\!\rightarrow\!\infty$ due to the first term $\alpha\bpsi^{1-\alpha}$, and likewise it diverges if any $\Psi_{ii}\!\rightarrow\! 0$ due to the second term $(1\!-\!\alpha) \bsigma \bpsi^{-\alpha}$. This proves the proposition for the case $\alpha\!\in\!(0,1)$.

%

Now suppose $\alpha\!>\!1$. Recall that the $\alpha$-divergence in eq.~(\ref{eq:renyi-gaussian}) is only defined for $\bphi_\alpha\!\succ\!0$ where $\bphi_\alpha = \alpha \bsigma^{-1} + (1\!-\!\alpha) \bpsi^{-1}$. But when $\alpha\!>\!1$, the condition $\bphi_\alpha\!\succ\!0$ can only be satisfied if $\Psi_{ii}\! >\! 0$ for all $i$. To see in addition that $\Psi_{ii}\! <\! \infty$, we proceed as before:
\begin{align} 
\argmin_{q \in \mathcal Q} \left [ \Ra(p||q) \right] & = \argmin_{\bpsi} \frac{1}{\alpha (\alpha - 1)} |\bsigma_\alpha|^{-\frac{1}{2}} |\bpsi|^\frac{\alpha}{2} \\
  &= \argmin_{\bpsi} \big|\bsigma\bphi_\alpha\bpsi\big|^{-\frac{1}{2}}\,\big|\bpsi\big|^{\frac{\alpha}{2}} \\
   &= \argmax_{\bpsi} \big|\bphi_\alpha\bpsi^{1-\alpha}\big|\label{eq:argmaxPhiPsi1} \\
  &= \argmax_{\bpsi} \big|\alpha\bsigma^{-1}\bpsi^{1-\alpha} + (1\!-\!\alpha)\bpsi^{-\alpha}\big|.\label{eq:argmaxPhiPsi2}
\end{align}
The determinant in eq.~(\ref{eq:argmaxPhiPsi1}) is strictly positive for any covariance~$\bpsi$ with bounded elements satisfying $\bphi_\alpha\!\succ\! 0$.
For $\alpha\!>\!1$ this same determinant, rewritten in eq.~(\ref{eq:argmaxPhiPsi2}), vanishes if any $\Psi_{ii}\!\rightarrow\!\infty$ because in this limit the matrices $\bpsi^{1-\alpha}$ and $\bpsi^{-\alpha}$ become low-rank. Thus the maximum must occur where $0\!<\!\Psi_{ii}\!<\infty$ for all $i$. This proves the proposition for $\alpha\!>\!1$.
\end{proof}

We can prove further properties of the variational approximation $q$ that minimizes $\Ra(q||p)$ in eq.~(\ref{eq:renyi-gaussian}). Though the minimizer of $\Ra(q||p)$ does not admit an analytical expression, it is possible to derive fixed-point equations that are satisfied by the variances~$\Psi_{ii}$ at this minimizer. As we shall see later, these fixed-point equations encode a great deal of information about variational approximations with the $\alpha$-divergence.

  \begin{mdframed}[hidealllines = true, backgroundcolor = Cerulean!10]
  \begin{proposition}[Fixed-point equations] \label{prop:renyi-solution}
   Let $\alpha \in \mathbb R^+ \backslash \{0, 1\}$, and let $p$ and~$q$ be given by eqs.~(\ref{eq:p_normal}--\ref{eq:q_normal}) where $\bpsi$ is diagonal. Then the $\alpha$-divergence in eq.~(\ref{eq:alpha-divergence}) is minimized when $\bnu=\bmu$ and the estimated precisions from $\bpsi$ satisfy the fixed-point equation
   \begin{equation} \label{eq:renyi-precision-matching}
    {\rm diag}\left(\bpsi^{-1}\right) = {\rm diag}\left(\bsigma_\alpha^{-1}\right),
  \end{equation}
or equivalently when the estimated variances from $\bpsi$ satisfy the fixed-point equation
  \begin{equation} \label{eq:varFixPtEq}
      {\rm diag}(\bpsi) = {\rm diag}\left(\bphi_\alpha^{-1}\right).
  \end{equation}
  \end{proposition} 
  \end{mdframed}

  \begin{proof}
  We derive the fixed-point equations by attempting to minimize the $\alpha$-divergence in eq.~(\ref{eq:renyi-gaussian}) with respect to the variational parameters. First, we consider the case where $\alpha\!\in\!(0,1)$. In this case, from eq.~(\ref{eq:argmin-alpha}), we have
\begin{equation}
\argmin_{q\in\mathcal{Q}}\Big[\Ra(p||q)\Big] = \argmin_{\bpsi}\Big[\log|\alpha\bpsi + (1\!-\!\alpha)\bsigma| - \alpha\log|\bpsi|\Big].
\label{eq:argmin_R_alpha}
\end{equation}
From the previous propositions, we know that the minimum of eq.~(\ref{eq:argmin_R_alpha}) occurs where \mbox{$0\!<\!\Psi_{ii}\!<\!\infty$}. 
Solving for the minimum, we find
\begin{align}
 0 &= \frac{\partial}{\partial \Psi_{ii}} [\log|\alpha\bpsi + (1\!-\!\alpha)\bsigma| - \alpha\log|\bpsi|\Big], \\ 
   &= \alpha [\alpha\bpsi + (1\!-\!\alpha)\bsigma]^{-1}_{ii} - \alpha\Psi^{-1}_{ii}, \\
   &= \alpha\left[\bsigma_\alpha^{-1}-\bpsi^{-1}\right]_{ii},
\end{align}
thus proving \cref{eq:renyi-precision-matching}. 

To derive the fixed-point equation in eq.~(\ref{eq:varFixPtEq}), we rewrite eq.~(\ref{eq:argmin_R_alpha}) as
  \begin{align}
\argmin_{q\in\mathcal{Q}}\Big[\Ra(p||q)\Big] 
  &= \argmin_{\bpsi}\Big[\log|\alpha\bpsi + (1\!-\!\alpha)\bsigma| - \alpha\log|\bpsi| + \log|\bpsi| - \log|\bpsi| \Big], \\
  &= \argmin_{\bpsi}\Big[\log\left|\alpha + (1\!-\!\alpha)\bsigma\bpsi^{-1}\right| -(1\!-\!\alpha)\log\left|\bpsi^{-1}\right|\Big], \\
  &= \argmin_{\bpsi^{-1}}\Big[\log\left|\alpha\bsigma^{-1} + (1\!-\!\alpha)\bpsi^{-1}\right| -(1\!-\!\alpha)\log\left|\bpsi^{-1}\right|\Big],
\end{align}
where in the last step we have exploited the fact that additive constants, such as $\log\left|\bsigma^{-1}\right|$, do not change the location of the minimum with respect to $\bpsi$ (or equivalently, with respect to~$\bpsi^{-1}$). Solving for the minimum, we find
\begin{equation}
 0 = \frac{\partial}{\partial \Psi^{-1}_{ii}} \Big[\log\left|\alpha\bsigma^{-1} + (1\!-\!\alpha)\bpsi^{-1}\right| -(1\!-\!\alpha)\log\left|\bpsi^{-1}\right|\Big] 
   = (1\!-\!\alpha)\left[\bphi_\alpha^{-1} - \bpsi\right]_{ii},
\end{equation}
which is equivalent to the claim that $\textrm{diag}(\bpsi) = \textrm{diag}\left(\bphi^{-1}_\alpha\right)$ in eq.~(\ref{eq:varFixPtEq}).

A similar procedure can be used to derive the fixed-point equations when $\alpha\! >\! 1$.
In this case, from eq.~(\ref{eq:argmaxPhiPsi2}) we have
\begin{equation} \label{eq:alpha-objective-alpha>1}
    \argmin_{q\in\mathcal{Q}}\Big[\Ra(p||q)\Big] =
      \argmax_{\bpsi}\big[ \log |\alpha \bpsi + (1\!-\!\alpha) \bsigma| - \alpha \log |\bpsi|\big].
\end{equation}
As before, from the previous propositions, we can exclude the edge cases $\Psi_{ii}\! =\! 0$ and $\Psi_{ii}\! =\! \infty$ as potential solutions, and therefore the maximum of the right side in eq.~(\ref{eq:alpha-objective-alpha>1}) is determined by the vanishing of its gradient with respect to $\bpsi$. But the gradients of eqs.~(\ref{eq:argmin_R_alpha}) and~(\ref{eq:alpha-objective-alpha>1}) are identical and therefore yield the same fixed-point equations.
\end{proof}

Proposition~\ref{prop:renyi-solution} provides two (equivalent) fixed-point equations for the variational parameter $\bpsi$, and it will prove useful to work with both in Section~\ref{sec:ordering}.
In componentwise notation, these fixed-point equations take the form
\begin{align}
    \label{eq:fixed-point-prec}
    \Psi^{-1}_{ii} &= \big[\alpha\bpsi + (1\!-\!\alpha)\bsigma\big]^{-1}_{ii}, \\
    \label{eq:fixed-point-var}
    \Psi_{ii} &= \left[\alpha\bsigma^{-1} + (1\!-\!\alpha)\bpsi^{-1}\right]^{-1}_{ii},  
\end{align}
and we see that eq.~(\ref{eq:fixed-point-prec}) reduces to the precision-matching solution $\Psi^{-1}_{ii}=\Sigma^{-1}_{ii}$ as $\alpha\rightarrow 0$ and eq.~(\ref{eq:fixed-point-var}) reduces to the variance-matching solution $\Psi_{ii}=\Sigma_{ii}$ as $\alpha\rightarrow 1$.



  %
  %

\subsection{Score-based Divergence}

Next we consider the score-based divergences presented in Table~\ref{tab:divergences}. These divergences are similar to the relative Fisher information~ \citep[eq. 7]{Courtade:2016} except that they use a \textit{weighted} norm~\citep[Appendix A]{Cai:2024} to measure the difference between the gradients of $\log p$ and $\log q$. When $p$ and $q$ are Gaussian, these score-based divergences are given by
\begin{align}
\label{eq:Sqp_weighted_norm}
S(q||p) &= \int \text d\bz\, q(\bz)\, \big\|\nabla \log q(\bz) - \nabla \log p(\bz) \big\|^2_{\text{Cov}(q)}, \\
\label{eq:Spq_weighted_norm}
S(p||q) &= \int \text d\bz\, q(\bz)\, \big\|\nabla \log q(\bz) - \nabla \log p(\bz) \big\|^2_{\text{Cov}(p)},
\end{align}
where we use $\|\mathbf{v}\|^2_\mathbf{A} = \mathbf{v}^\top\mathbf{A}\mathbf{v}$ to denote the weighted norm for any $\mathbf{A}\!\succ\!\mathbf{0}$. The score-based divergences in eqs.~(\ref{eq:Sqp_weighted_norm}-\ref{eq:Spq_weighted_norm}) are dimensionless measures of the discrepancy between $p$ and~$q$, and in particular, like the KL and $\alpha$-divergences in the previous sections, they are invariant to affine reparameterizations of the support of these distributions \cite[Theorem A.4]{Cai:2024}. For $p$ and $q$ in eqs.~(\ref{eq:p_normal}--\ref{eq:q_normal}), these divergences are given by
\begin{align}
\label{eq:Sqp}
S(q||p) &= \text{tr}\left[\left(\mathbf{I}\!-\!\bpsi\bsigma^{-1}\right)^2\right] + 
  (\bnu\!-\!\bmu)^\top \bsigma^{-1}\bpsi\bsigma^{-1}(\bnu\!-\!\bmu), \\
\label{eq:Spq}
S(p||q) &= \text{tr}\left[\left(\mathbf{I}\!-\!\bsigma\bpsi^{-1}\right)^2\right] + 
  (\bmu\!-\!\bnu)^\top \bpsi^{-1}\bsigma\bpsi^{-1}(\bmu\!-\!\bnu),
\end{align}
as also shown in \citet{Cai:2024}. The rest of this section examines the solutions that minimize these expressions with respect to the variational mean $\bnu$ and diagonal covariance~$\bpsi$.

First we consider the minimization of the \textit{reverse} score-based divergence, $S(q||p)$. This minimization can be formulated as a nonnegative quadratic program (NQP)---that is, a convex instance of quadratic programming with nonnegativity constraints.
  \begin{mdframed}[hidealllines = true, backgroundcolor = Cerulean!10]
  \begin{proposition}[NQP for minimizing $S(q||p)$]  \label{prop:solution-fisher-qp}
    Let $p$ and $q$ be given by eqs.~(\ref{eq:p_normal}--\ref{eq:q_normal}) where $\bpsi$ is diagonal. Then the reverse score-based divergence in eq.~(\ref{eq:Sqp}) is minimized by setting $\bnu\!=\!\bmu$ and solving the quadratic program
    \begin{equation}  \label{eq:solution-fisher-qp}
     \min_{\bs\geq 0} \left[\tfrac{1}{2} \bs^\top \bH\, \bs - \mathbf{1\!}^\top\!\bs\right],
    \end{equation}
    where $\bs$ is constrained to lie in the nonnegative orthant, $\mathbf{1}$ is the vector of all ones, 
    $\bH$ is the positive-definite matrix with elements
    \begin{equation}
      H_{ij} = \frac{\left(\Sigma^{-1}_{ij}\right)^2}{\Sigma^{-1}_{ii} \Sigma^{-1}_{jj}},
      \label{eq:H}
    \end{equation}
    and $\bpsi$ is obtained from the solution of eq.~(\ref{eq:solution-fisher-qp}) by identifying $s_i = \Psi_{ii} \Sigma^{-1}_{ii}$.
  \end{proposition}
  \end{mdframed}
  \vspace{1ex}
  
  \begin{proof}
  It is clear that eq.~(\ref{eq:Sqp}) is minimized by setting $\bnu\!=\!\bmu$ in the rightmost term. 
  For the remaining term, we note that
  \begin{equation}
  \tfrac{1}{2}\,\text{tr}\left[\left(\mathbf{I}\!-\!\bpsi\bsigma^{-1}\right)^2\right]\
    =\ \tfrac{n}{2}\, -\, \sum_i \Psi_{ii}\Sigma_{ii}^{-1}\, +\,
        \tfrac{1}{2}\sum_{ij} \Psi_{ii}\Psi_{jj}\left(\Sigma^{-1}_{ij}\right)^2,
\label{eq:Sqp_tr}
 \end{equation}
and the NQP in eq.~(\ref{eq:solution-fisher-qp}) is obtained by defining $s_i = \Psi_{ii}\Sigma_{ii}^{-1}$. Finally, we observe that the matrix $\mathbf{H}$ in eq.~(\ref{eq:H}) is always positive-definite. Indeed, $\mathbf{H}$ is the Hadamard (i.e., component-wise) square of the correlation matrix built from the precision matrix $\mathbf{\Sigma}^{-1}$, and we note 
from the Schur product theorem 
that the Hadamard product of two positive-definite matrices is also positive-definite~\citep{Horn:2012}.
\end{proof}

Given the symmetry of eqs.~(\ref{eq:Sqp}-\ref{eq:Spq}), it is not surprising that we obtain an analogous optimization when minimizing the \textit{forward} score-based divergence, $S(p||q)$, in terms of the variational parameters $\bnu$ and $\bpsi$. In particular, we have the following result.
  \begin{mdframed}[hidealllines = true, backgroundcolor = Cerulean!10]
  \begin{proposition}[NQP for minimizing $S(q||p)$]  \label{prop:solution-fisher-pq}
    Let $p$ and $q$ be given by eqs.~(\ref{eq:p_normal}--\ref{eq:q_normal}) where $\bpsi$ is diagonal. Then the forward score-based divergence in eq.~(\ref{eq:Sqp}) is minimized by setting $\bnu\!=\!\bmu$ and solving the quadratic program
    \begin{equation}  \label{eq:solution-fisher-pq}
     \min_{\bt\geq 0} \left[\tfrac{1}{2} \bt^\top \mathbf{J}\, \bt - \mathbf{1\!}^\top\!\bt\right],
    \end{equation}
    where $\bt$ is constrained to lie in the nonnegative orthant, 
    $\mathbf{J}$ is the positive-definite matrix with elements
    \begin{equation}
      J_{ij} = \frac{\left(\Sigma_{ij}\right)^2}{\Sigma_{ii} \Sigma_{jj}},
      \label{eq:J}
    \end{equation}
    and $\bpsi$ is obtained from the solution of eq.~(\ref{eq:solution-fisher-pq}) by identifying $t_i = \Psi^{-1}_{ii} \Sigma_{ii}$.
  \end{proposition}
  \end{mdframed}
\vspace{1ex}

\begin{proof}
The proof follows the same steps as in the previous proposition, but with the matrices~$\bpsi$ and $\bsigma$ playing the reverse roles as they did in eq.~(\ref{eq:Sqp_tr}).
\end{proof}

We conclude this section by noting a peculiar property of the score-based divergences in \Cref{tab:divergences}. It is possible for the solutions of the NQPs in Propositions~\ref{prop:solution-fisher-qp} and \ref{prop:solution-fisher-pq} to lie on the boundary of the nonnegative orthant. That is, there may exist some $i^\text{th}$ component along the diagonal of $\bpsi$ such that $S(q||p)$ is minimized by setting $\Psi_{ii}\!=\!0$ (in Proposition~\ref{prop:solution-fisher-qp}) or such that $S(p||q)$ is minimized by setting $\Psi_{ii}^{-1}\!=\!0$ (in Proposition~\ref{prop:solution-fisher-pq}). Strictly speaking, \textit{solutions of this form---with zero or infinite marginal variances---do not define proper distributions that lie in the family $\mathcal{Q}$ of factorized Gaussian approximations}. Such solutions do not arise with FG-VI based on the KL or $\alpha$-divergences. This phenomenon gives rise to the following definition.

  \begin{mdframed}[hidealllines = true, backgroundcolor = Cerulean!10]
  \begin{definition} \label{def:variational-collapse}
    Let $\mathcal{Q}$ denote the family of factorized Gaussian approximations in eq.~(\ref{eq:q_factor}) that are proper distributions (i.e., with $0\!<\!\mathcal{H}(q)\!<\!\infty$). For some target distribution $p$ and divergence $D$, we say that FG-VI undergoes \textbf{variational collapse} when $\arg\!\inf_{q\in\mathcal{Q}} D(q,p) \not\in \mathcal{Q}$.
  \end{definition}
  \end{mdframed}
\vspace{1ex}

\begin{figure}[t]
    \begin{center}
    \includegraphics[width=\textwidth]{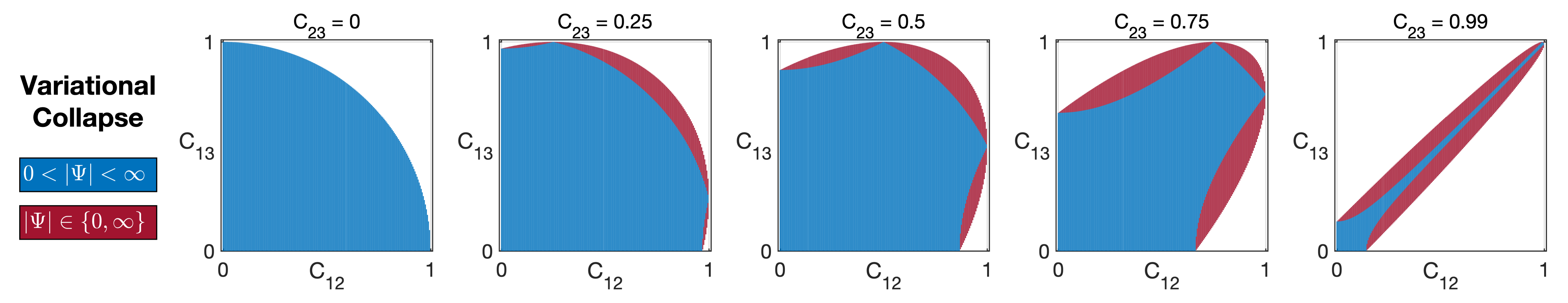}
    \end{center}
    \caption{{When FG-VI is based on minimizing the score-based divergences in Table~\ref{tab:divergences}, it may estimate zero or infinite values for the marginal variances. The red areas indicate these occurrences of \textit{variational collapse} when FG-VI with a score-based divergence is used to approximate a three-dimensional Gaussian with a non-diagonal correlation matrix $\mathbf{C}$.}}
    \label{fig:variational-collapse}
\end{figure}

Fig.~\ref{fig:variational-collapse} illustrates this collapse when FG-VI is used to approximate a multivariate Gaussian distribution in three dimensions with varying degrees of correlation; the amount of correlation is determined by the magnitude of the off-diagonal elements $C_{12}$, $C_{23}$, and~$C_{13}$ in eq.~(\ref{eq:correlation}). These off-diagonal elements are constrained by the fact that the correlation matrix as a whole must be positive definite. Each panel in the figure visualizes a two-dimensional slice of the three-dimensional (convex) set of positive-definite correlation matrices for a fixed value of $C_{23}$. The blue regions of each slice indicate the areas where the NQPs in Propositions~\ref{prop:solution-fisher-qp} and \ref{prop:solution-fisher-pq} yield finite and nonzero estimates of the variances; the red regions indicate areas of variational collapse. (It can be shown, in three dimensions, that the forward and reverse score-based divergences share the same areas of variational collapse.) Interestingly, the leftmost panel shows that variational collapse does not occur in two dimensions or arise purely from pairwise correlations. On the other hand, the remaining panels show that variational collapse occurs in three dimensions whenever the correlation matrix is dense and has at least one off-diagonal element of sufficiently large magnitude.

The above example shows that variational collapse occurs when some pairwise correlations are large and the assumption of factorization is strongly violated. This may not be a problem in settings where we expect the target covariance $\bsigma$ to be sparse.
However, in many settings, we do not have a priori knowledge on the structure of~$\bsigma$.
In this case, it seems fraught to minimize a score-based divergence, particularly if the goal of FG-VI is to provide some quantification of uncertainty
(whether it be with marginal variances, marginal precisions or generalized variance).
Still, a score-based divergence may work well in conjunction with a richer family of approximations---for instance, a multivariate Gaussian with a dense covariance matrix, as demonstrated by~\citet{Cai:2024}.

\section{Ordering of Divergences for FG-VI} \label{sec:ordering}

We have seen that different divergences $D(q,p)$ yield 
different solutions to the problem of variational inference in eq.~(\ref{eq:argminDqp}); we have also seen,
in turn, that these different solutions yield different estimates of the marginal variances, precisions, and entropy.
In the setting where both $p$ and $q$ are Gaussian, these estimates provide a natural way to compare and even order the divergences; see Definition~\ref{def:ordering}.
In this section, we show that the divergences in Table~\ref{tab:divergences} can be ordered in this way; that is, for $0\!<\!\alpha_1\!<\!\alpha_2\!<\! 1$, we prove that
\begin{equation*}
  S(q||p) \prec \KL(q||p) \prec \text{D}_{\alpha_1}(p||q) \prec \text{D}_{\alpha_2}(p||q) \prec \KL(p||q) \prec S(p||q),
\end{equation*}
as stated in \Cref{thm:ordering}.
We also provide slightly weaker results on $\alpha$-divergences for the case were $\alpha > 1$.
\begin{figure}[t]
    \includegraphics[width=1.8in]{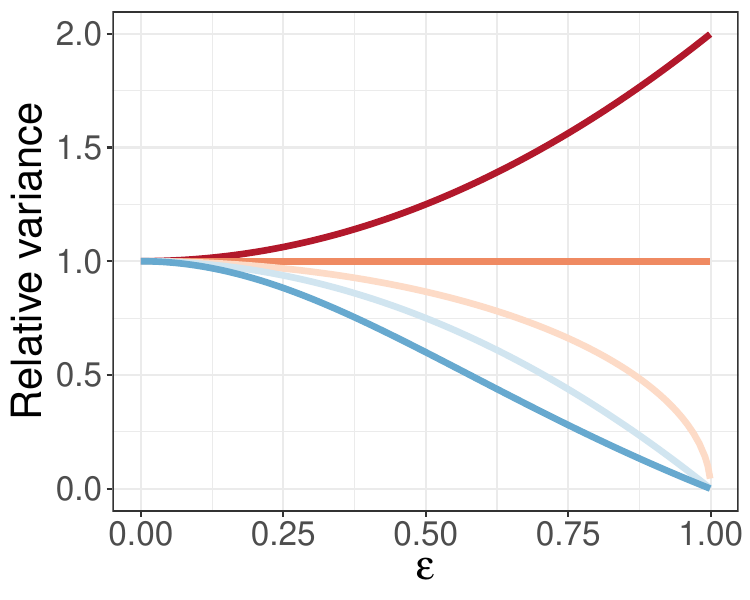} \ 
    \includegraphics[width=1.8in]{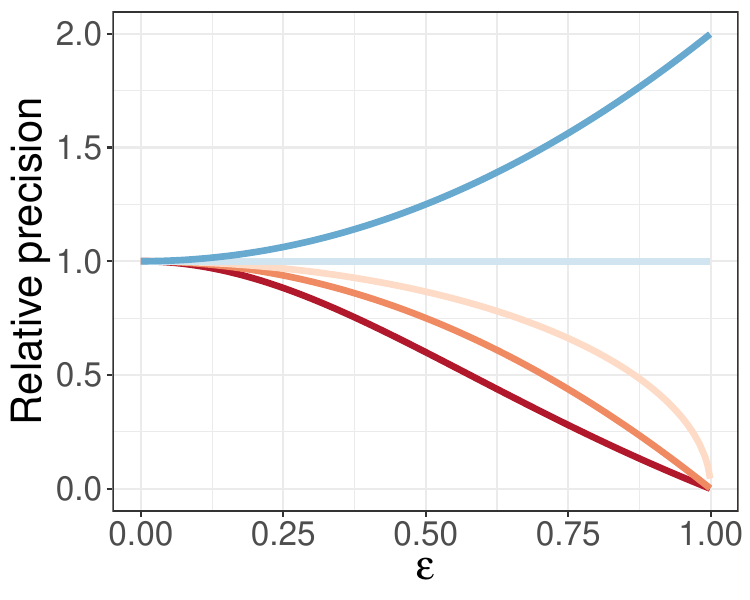} \ 
    \includegraphics[width=2.1in]{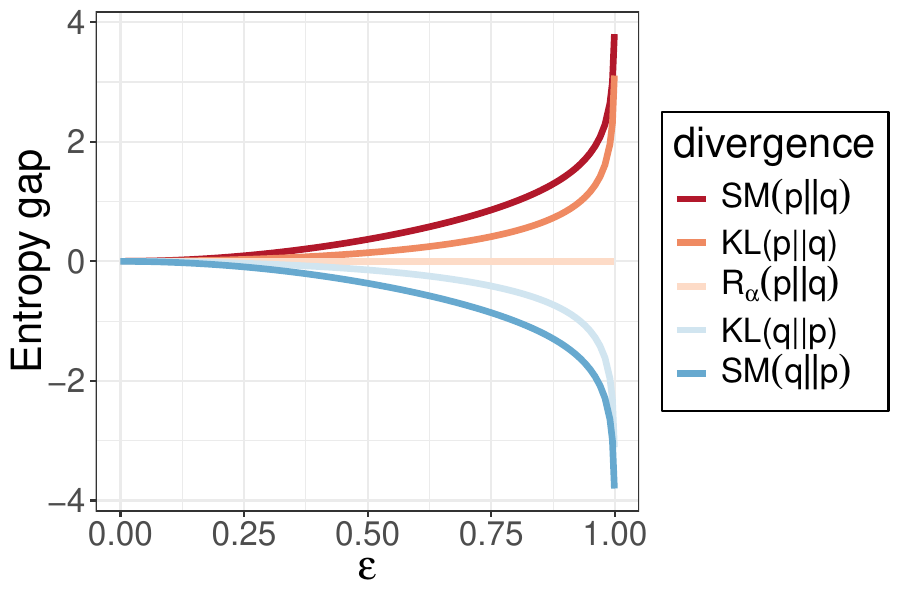}
    \caption{{Variances, precisions, and entropy estimated by FG-VI with different divergences. In the left and center panels, the variance is normalized by the variance of $p$, and the precision by the precision of $p$, both along the first coordinate. In the right panel, we plot the difference between the estimated entropy and the entropy of $p$.    
    Here FG-VI was used to approximate a 2-dimensional Gaussian with correlation~$\varepsilon$. The $\alpha$-divergence was computed for $\alpha\!=\!0.5$. The ordering of the curves matches the predictions of \Cref{thm:ordering}.}}
    \label{fig:ordering-uncertainty}
\end{figure}

Figure~\ref{fig:ordering-uncertainty} illustrates the consequences of this ordering when FG-VI is used to approximate a 2-dimensional Gaussian target with varying degrees of correlation. As the amount of correlation (denoted by $\varepsilon$) increases from zero to one---from no correlation to perfect correlation---the target Gaussian more starkly violates the assumption of factorization. From left to right, the panels plot the estimates of variance, precision, and entropy that are obtained from FG-VI with different divergences, and we see that for all values of~$\varepsilon$, these estimates are ordered exactly as predicted by \Cref{thm:ordering}.

We have already proven one of these orderings. Recall from section~\ref{sec:KL} that when FG-VI is based on minimizing $\KL(q||p)$, it underestimates the marginal variances, at least one of them strictly.
On the other hand, when FG-VI is based on minimizing $\KL(p||q)$, it correctly estimates the marginal variances. Thus we have already shown that $\KL(q||p)\prec \KL(q||p)$. 
The rest of this section is devoted to proving the other orderings in \Cref{thm:ordering}.


%


\subsection{Ordering of Score-based Divergences} \label{sec:ordering-score}

In this section we prove the two outermost orderings in \Cref{thm:ordering}; specifically, we show that \mbox{$S(q||p)\!\prec\!\KL(q||p)$} and $\KL(p||q)\!\prec\! S(p||q)$. At a high level, these proofs are obtained by analyzing the Karush-Kuhn-Tucker (KKT) conditions of the quadratic programs in Propositions~\ref{prop:solution-fisher-qp} and \ref{prop:solution-fisher-pq}.


 \vspace{2ex}
 \noindent
 \fcolorbox{Cerulean!10}{Cerulean!10}{\textit{Proof of $S(q||p)\prec \KL(q||p)$.}}
   Consider the diagonal covariance matrix $\bpsi$ that FG-VI estimates by minimizing $S(q||p)$ on the left side of this ordering. This matrix is obtained from the solution of the NQP in \cref{eq:solution-fisher-qp} for the unknown nonnegative variables \mbox{$s_i = \Psi_{ii}\Sigma^{-1}_{ii}$}. The solution to this NQP must satisfy the KKT conditions: namely, for each component of~$\mathbf{s}$, we have either that (i) $s_i\!=\!0$ and $(\mathbf{Hs})_i>\!1$ or (ii) $s_i\!>\!0$ and $(\mathbf{Hs})_i\!=\!1$. We examine each of these cases in turn:

    \vspace{1ex}
    \begin{center}
    \begin{tabular}{rl}
    \fbox{$s_i\!=\!0$} & 
      \begin{minipage}[t]{5in}
       This is a case of variational collapse where by minimizing $S(q||p)$ we estimate that $\Psi_{ii}=0$.
      \end{minipage} \\[4ex]
   \fbox{$s_i\!>\!0$} & 
      \begin{minipage}[t]{5in}
       In this case we observe that $(\mathbf{Hs})_i\!=\!1$ from the KKT condition and also that $H_{ij}\geq 0$ and $H_{ii}=1$ from eq.~(\ref{eq:H}). Thus we see that
    \begin{equation}
    \label{eq:slt1}
      s_i = H_{ii} s_i = 1 - \sum_{j\neq i}H_{ij} s_j \leq 1,\\[-2ex]
    \end{equation}
       or equivalently, that $\Psi_{ii}\Sigma^{-1}_{ii} \leq 1$.
      \end{minipage}
   \end{tabular}
\end{center}
    
\noindent 
We have thus shown that $\Psi_{ii}\Sigma_{ii}^{-1}\!\leq\! 1$ for all of the marginal variances estimated by minimizing the score-based divergence $S(q||p)$. 
Next we show that this inequality must be strict for at least one component. The claim is trivially true if any $s_i\!=\!0$, since this implies $\Psi_{ii}\Sigma_{ii}^{-1}\!=\!0$. Suppose to the contrary that every element of $\mathbf{s}$ is strictly positive. Since by assumption $\bsigma$ is not diagonal, it must also be true that $\bsigma^{-1}$ is not diagonal, and therefore from eq.~(\ref{eq:H}) we can pick some~$i$ and~$k$ such that $H_{ik}\!>\!0$.  But then, continuing from eq.~(\ref{eq:slt1}), we see that
\begin{equation}
    \label{eq:slt2}
    s_i = 1-\sum_{j\neq i} H_{ij} s_j \leq 1-H_{ik}s_k < 1,
\end{equation}
or equivalently that $\Psi_{ii}\Sigma_{ii}^{-1}\!<\!1$, thus proving the claim. In sum we have shown that \mbox{$\Psi_{ii}\!\leq\! 1 / \Sigma^{-1}_{ii}$} for all of the marginal variances estimated by minimizing $S(q||p)$, and also that this inequality is strict for at least one component. Finally, we recall from eq.~(\ref{eq:precision-matching}) that $1 / \Sigma^{-1}_{ii}$ is the marginal variance estimated by minimizing $\KL(q||p)$. Per Definition~\ref{def:ordering} we conclude that $S(q||p)\prec\KL(q||p)$.
\mbox{\hspace{55ex}$\blacksquare$}
\vspace{1ex}

  We use the same methods to prove an analogous ordering for the forward score-matching and KL divergences.


 \vspace{2ex}
 \noindent
 \fcolorbox{Cerulean!10}{Cerulean!10}{\textit{Proof of $S(p||q)\succ \KL(p||q)$.}}
 To prove the order we consider the variances $\Psi_{ii}$ that minimize $S(q||p)$; they are found by solving the NQP in \cref{eq:solution-fisher-pq} for the nonnegative variables \mbox{$t_i = \Psi_{ii}^{-1}\Sigma_{ii}$}. The solution must satisfy the KKT conditions: either (i) $t_i\!=\!0$ and $(\mathbf{Jt})_i>\!1$ or (ii) $t_i\!>\!0$ and $(\mathbf{Jt})_i\!=\!1$. We examine each of these cases in turn: 

     \vspace{1ex}
    \begin{center}
    \begin{tabular}{rl}
    \fbox{$t_i\!=\!0$} & 
      \begin{minipage}[t]{5in}
       This is a case of variational collapse where by minimizing $S(p||q)$ we estimate that $\Psi^{-1}_{ii}\!=\!0$; that is, $\Psi_{ii}\! =\! \infty$, and so $\Psi_{ii}\! >\! \Sigma_{ii}$. 
      \end{minipage} \\[3ex]
    \end{tabular}
    \end{center}
    \begin{center}
     \begin{tabular}{rl}
   \fbox{$t_i\!>\!0$} & 
      \begin{minipage}[t]{5in}
       In this case, we observe that $(\mathbf{Jt})_i=1$ from the KKT condition and also that $J_{ij}\!\geq\!0$ and $J_{ii}\!=\!1$ from eq.~(\ref{eq:J}). Thus we see that
    \begin{equation}
      t_i = J_{ii}t_i = 1 - \sum_{j\neq i}J_{ij}t_j \leq 1,
    \end{equation}
    or equivalently, that $\Psi_{ii}^{-1}\Sigma_{ii}\!\leq\! 1$ and $\Psi_{ii}\!\geq\!\Sigma_{ii}$. Moreover, since $\bsigma$ is non-diagonal, ${\bf J}$ must also be non-diagonal, and hence there must exist at least one coordinate $j$ such that $\Psi_{jj}\! >\! \Sigma_{jj}$.
      \end{minipage}
   \end{tabular}
\end{center}
\vspace{1ex}
\noindent
We have shown that $\Psi_{ii}\!\geq\!\Sigma_{ii}$ for all $i$, and also that this inequality is strict for at least one component of the variance. Finally, we recall that the correct marginal variance (namely,~$\Sigma_{ii}$) is estimated by VI when minimizing $\KL(p||q)$. Per Definition~\ref{def:ordering} we conclude that $S(p||q)\!\succ\! \KL(p||q)$.~\mbox{\hspace{66ex}$\blacksquare$}
 

\subsection{Ordering of KL and $\alpha$-divergences}
\label{sec:ordering-KL}

Next we prove the two intermediate orderings in \Cref{thm:ordering}; specifically, for any $\alpha\!\in\!(0,1)$, we show that $\KL(q||p)\prec\Ra(p||q) \prec\KL(p||q)$.
These proofs are obtained by applying the inequalities in Lemma~\ref{prop:hadamard} to the fixed-point equations for the optimal marginal variances in eqs.~(\ref{eq:fixed-point-prec}--\ref{eq:fixed-point-var}), obtained by minimizing the $\alpha$-divergence.
The same approach allows us to show that $\Ra(p||q) \succ \KL(p||q)$ for $\alpha\! >\! 1$.

\vspace{2ex}
\noindent
\fcolorbox{Cerulean!10}{Cerulean!10}{\textit{Proof of $\KL(q||p)\prec \Ra(p||q)$ for $\alpha\!\in\!(0,1)$.}}
 Let $\bpsi$ denote the diagonal covariance matrix estimated by minimizing $\Ra(p||q)$, and consider the matrix $[\alpha\bsigma^{-1} + (1\!-\!\alpha)\bpsi]^{-1}$ that appears on the right side of its fixed-point equation in eq.~(\ref{eq:fixed-point-var}). 
 For $\alpha\!\in\!(0,1)$, it is clear that this matrix is positive-definite and non-diagonal, and thus we can apply the inequality in eq.~(\ref{eq:Ageq}) of Lemma~\ref{prop:hadamard} to its diagonal elements. In this way, we find:
 \begin{align}
 \alpha\Psi_{ii}\Sigma^{-1}_{ii}
   &= \Psi_{ii}\left[\alpha\bsigma^{-1}+(1\!-\!\alpha)\bpsi^{-1}\right]_{ii}\, -\, (1\!-\!\alpha) \\
   &\geq \frac{\Psi_{ii}}{\left[\alpha\bsigma^{-1}+(1\!-\!\alpha)\bpsi^{-1}\right]^{-1}_{ii}}\, -\, (1\!-\!\alpha), \\
   &= (\Psi_{ii}/\Psi_{ii})\, -\, (1\!-\!\alpha), \\
   &= \alpha.
 \end{align}
 Dividing both sides by $\alpha\Sigma^{-1}_{ii}$, we see that $\Psi_{ii}\geq1/\Sigma_{ii}^{-1}$ for all $i$. Next we appeal to the strict inequality of Lemma~\ref{prop:hadamard} and conclude that $\Psi_{jj}>1/ \Sigma_{jj}^{-1}$ for some $j$. On the left and right sides of these inequalities appear, respectively, the marginal variances estimated by minimizing $D_\alpha(p,q)$ and $\KL(q,p)$. Thus we have shown $\Ra(p||q)\succ \KL(q,p)$.\mbox{\hspace{25ex}$\blacksquare$}

\vspace{2ex}
\noindent
\fcolorbox{Cerulean!10}{Cerulean!10}{\textit{Proof of $\Ra(p||q)\prec \KL(p||q)$ for $\alpha\!\in\!(0,1)$.}}
The proof follows the same steps as the previous one, but now we combine the inequalities of Lemma~\ref{prop:hadamard} with the fixed-point equation in eq.~(\ref{eq:fixed-point-prec}). In this way we find:
\begin{equation}
 (1\!-\!\alpha)\Psi^{-1}_{ii}\Sigma_{ii}\,
   =\, \Psi^{-1}_{ii}\left[\alpha\bpsi+(1\!-\!\alpha)\bsigma\right]_{ii}\! -\! \alpha\,
   \geq\, \frac{\Psi^{-1}_{ii}}{\left[\alpha\bpsi+(1\!-\!\alpha)\bsigma\right]^{-1}_{ii}}\! -\! \alpha\,
   =\, 1\!-\!\alpha.
   \label{eq:orderingFromLemma}
\end{equation}
 Dividing both sides by $(1\!-\!\alpha)\Psi^{-1}_{ii}$, we see that $\Sigma_{ii}\geq\Psi_{ii}$ for all $i$, and we know from the strict inequality in eq.~(\ref{eq:Agt}) of Lemma~\ref{prop:hadamard} that $\Sigma_{jj}>\Psi_{jj}$ for some $j$. On the left and right sides of these inequalities appear, respectively, the marginal variances estimated by minimizing $\KL(p||q)$ and $\Ra(p||q)$. Thus we have shown $\KL(p||q)\!\succ\!\Ra(p||q)$ for $\alpha\!\in\!(0,1)$.~\mbox{\hspace{11ex}$\blacksquare$}

\vspace{2ex}
\noindent
\fcolorbox{Cerulean!10}{Cerulean!10}{\textit{Proof of $\Ra(p||q)\succ \KL(p||q)$ for $\alpha\!>\!1$.}}
To prove this ordering we need to justify the steps in eq.~(\ref{eq:orderingFromLemma}) when $\alpha\!>\!1$, as opposed to when $\alpha\!\in\!(0,1)$. But when $\alpha\!>\!1$, it is not obvious that the matrix $\bsigma_\alpha = \alpha\bpsi + (1\!-\!\alpha)\bsigma$ is positive-definite; this must be demonstrated before we can apply the inequality in Lemma~\ref{prop:hadamard}. Therefore we begin by showing that $\bsigma_\alpha$ is positive-definite whenever the matrices $\bpsi$, $\bpsi^{-1}$, and $\bphi_\alpha$ are positive-definite. (We know that $\bpsi\!\succ\! 0$ and $\bpsi^{-1}\!\succ\!0$ from Proposition \ref{prop:variance-bounds}, and we know that $\bphi_\alpha\!\succ\!0$ because it is necessary for the $\alpha$-divergence between two multivariate Gaussians to be well-defined.) Recall that $\bsigma_\alpha = \bsigma \bphi_\alpha \bpsi$. It follows that
\begin{equation}
  \bpsi^{-\frac{1}{2}}\bsigma_\alpha\bpsi^{-\frac{1}{2}}
    = \left(\bpsi^{-\frac{1}{2}}\bsigma^{\frac{1}{2}}\right)
      \left(\bsigma^\frac{1}{2}\bphi_\alpha\bsigma^\frac{1}{2}\right)
      \left(\bpsi^{-\frac{1}{2}}\bsigma^{\frac{1}{2}}\right)^{-1}.
\end{equation}
Thus the matrix $\bpsi^{-\frac{1}{2}}\bsigma_\alpha\bpsi^{-\frac{1}{2}}$ on the left side and the inner matrix $\bsigma^\frac{1}{2}\bphi_\alpha\bsigma^\frac{1}{2}$ on the right side are related by a \textit{similarity transformation} and therefore share the same eigenvalues.\footnote{For a matrix ${\bf A} \in \mathbb R^{n \times n}$ and an invertible matrix ${\bf B} \in \mathbb R^{n \times n}$, it is well known and straightforward to show that the matrices ${\bf A}$ and ${\bf BAB}^{-1}$ have the same eigenvalues.}
Since $\bsigma^\frac{1}{2}\bphi_\alpha\bsigma^\frac{1}{2}$ is positive-definite, the eigenvalues shared by these two matrices are positive.
Next, we note that the matrix $\bpsi^{-\frac{1}{2}}\bsigma_\alpha\bpsi^{-\frac{1}{2}}$ is symmetric, and since all of its eigenvalues are positive, it is also positive-definite.
Finally, we can write
\begin{equation}
    {\bsigma}_\alpha = \bpsi^\frac{1}{2} \left (\bpsi^{-\frac{1}{2}} \bsigma_\alpha \bpsi^{-\frac{1}{2}} \right) \bpsi^\frac{1}{2},
\end{equation}
and since the inner matrix $\bpsi^{-\frac{1}{2}}\bsigma_\alpha\bpsi^{-\frac{1}{2}}$ on the right side is positive-definite, it follows that~$\bsigma_\alpha$ is also positive-definite.
We can now apply Lemma~\ref{prop:hadamard} exactly as in eq.~(\ref{eq:orderingFromLemma}) to obtain
\begin{equation}
    (1-\alpha)\Psi_{ii}^{-1} \Sigma_{ii} \ge 1 - \alpha
\end{equation}
with a strict inequality for some $j$. But now, because $\alpha\!>\!1$, we obtain the \textit{reverse} inequality when dividing both sides by $(1\!-\!\alpha)$, and we conclude that
$\Psi_{ii} \ge \Sigma_{ii}$ with a strict inequality for some $j$. Thus we have shown $\Ra(p||q)\!\succ\!\KL(p||q)$ for $\alpha\!>\!1$.  \mbox{\hspace{27ex}$\blacksquare$}


\subsection{Ordering of $\alpha$-divergences for $\alpha \in (0, 1)$} \label{sec:ordering-alpha}

In this section we prove the ordering of the $\alpha$-divergences in \Cref{thm:ordering}. This proof is more technical, relying on a detailed analysis of the fixed point equations in Proposition~\ref{prop:renyi-solution}.

\vspace{2ex}
\noindent
\fcolorbox{Cerulean!10}{Cerulean!10}{\textit{Proof of $\text{D}_{\alpha_1}(p||q)\prec \text{D}_{\alpha_2}(p||q)$ for $0\!<\alpha_1\!<\!\alpha_2\!<\!1$.}} 
Let $\bpsi(\alpha)$ denote the diagonal covariance matrix that FG-VI estimates by minimizing $\Ra(p||q)$ and note by inspection of $\Ra(p||q)$ that $\bpsi(\alpha)$ is smooth.
%
%
For $\alpha\!\in\!(0,1)$, we know from the results of \Cref{sec:ordering-KL} that
$\Psi_{ii}(\alpha)$ is sandwiched between its limiting values at zero and~one:
\begin{equation}
   \frac{1}{\Sigma_{ii}^{-1}}\, =\,
   \lim_{\alpha\rightarrow 0^+} \Psi_{ii}(\alpha)\, \leq\, 
   \Psi_{ii}(\alpha)\, \leq\,
   \lim_{\alpha\rightarrow 1^-} \Psi_{ii}(\alpha)\, =\,
   \Sigma_{ii}.
\end{equation}
We also know from Lemma~\ref{prop:hadamard} that $\Sigma_{ii}\Sigma^{-1}_{ii}>1$ whenever $\sum_{j\neq i}|\Sigma_{ij}|>0$, and that in this case, the above inequalities are strict.
A useful picture of this situation is shown in the left panel of Figure~\ref{fig:convex-function}. Consider any component $\Psi_{ii}(\alpha)$ of the estimated variances that is \textit{not} constant on the unit interval. (As shown previously, there must be at least one such component.) In this case, there are only two possibilities: either $\Psi_{ii}(\alpha)$ is strictly increasing, or it is not. If the former is always true, then it follows that $\text{D}_{\alpha_1}(p||q)\prec \text{D}_{\alpha_2}(p||q)$ whenever $0\!<\!\alpha_1\!<\!\alpha_2\!<\!1$. We shall prove \textit{by contradiction} that this is indeed the case.

Assume that there are one or more diagonal components, $\Psi_{ii}(\alpha)$, that are neither constant nor strictly increasing over the unit interval; also, let $\mathcal{I}$ denote the set that contains the indices of these components. Since each such component must have at least one stationary point, we can also write:
\begin{equation}
\mathcal{I} = \Big\{i 
  \,\Big|\,\Sigma_{ii}\Sigma_{ii}^{-1}\!>\!1\ \mbox{and}\ \Psi_{ii}'(\alpha)=0\ \mbox{for some}\ \alpha\in(0,1)\Big\}.
\end{equation}
For each such component, there must also exist some \textit{minimal} point $\tau_i\in(0,1)$ where its derivative vanishes. We define $\tau = \min_{i\in\mathcal{I}} \tau_i$ and $j=\arg\!\min_{i\in\mathcal{I}} \tau_i$, so that by definition
\begin{align}
 \Psi_{ii}'(\tau) &\geq 0\quad\mbox{for all $i$}, \label{eq:tau-psd}\\
 \Psi_{jj}'(\tau) &= 0. \label{eq:dPsijjtau}
\end{align}
To obtain the desired contradiction, we shall prove that there exists no point $\tau$ (as imagined in Figure~\ref{fig:convex-function}) with these properties.

\begin{figure}
  \centering
    \begin{tikzpicture}[scale=3.5]
    \draw[->] (0,0) -- (1.1,0) node[right] {$\alpha$};
    \draw[->] (0,0) -- (0,1.1) node[above] {\textcolor{MidnightBlue}{$\Psi_{jj}(\alpha)$}};

    \draw[Orange, thick, domain=0:1, samples=100] plot (\x, {0.25 * sin(0.035 * pi * deg(\x) * deg(\x)) + 0.8 * \x + 0.2} );

    \draw[MidnightBlue, thick, domain=0:1, samples=100] plot (\x, {0.7 * \x^2 + 0.1 * \x + 0.2} );

    \draw[black, dashed, domain=0:1, samples=100] plot (\x, 0.2);
    \draw[black, dashed, domain=0:1, samples=100] plot (\x, 1);
    
    \draw[black, fill] (0, 0.2) circle (0.25pt) node[left] {$1 / \Sigma^{-1}_{jj}$};
    \draw[black, fill] (0, 1) circle (0.25pt) node[left] {$\Sigma_{jj}$};
    \draw[black, fill] (1, 1) circle (0.25pt);

    \draw[black, fill] (1, 0) circle (0.5pt) node[below] {1};
    \draw[black, fill] (0, 0) circle (0.5pt) node[below] {0};
    \draw[orange, fill] (0.55, 0) circle (0.5pt) node[below] {$\textcolor{orange}{\tau}$};
    \draw[orange, dotted] (0.55, 0) -- (0.55, 0.85);

  \end{tikzpicture}
  \begin{tikzpicture}[scale=3.5]
    \draw[->] (0,0) -- (1.1,0) node[right] {$\alpha$};
    \draw[->] (0,0) -- (0,1.1) node[above] {$f(\alpha)$};
    
    \draw[black, thick, domain=0:1, samples=100] plot (\x, {(\x + 0.22)^2 - \x + 0.5});

    \draw[black, thick, dashed, domain=-0.25:0.25, samples=100] plot (\x, -0.6 * \x + 0.22^2 + 0.5);
    \draw[black, thick, dashed, domain=-0.3:0.2, samples=100] plot (\x + 0.7,  0.7 * \x + 0.635);

    \draw[orange, thick, dashed, domain=0:0.55, samples=100] plot (\x, 0.22^2 + 0.5);

    \draw[orange, fill] (0, 0.22^2 + 0.5) circle (0.5pt);
   \draw[orange, fill] (0.55, 0.22^2 + 0.5) circle (0.5pt) node[right] {\hspace{1ex}$ \textcolor{orange}{f(0) = f(\tau) = \Psi_{jj}(\tau)}$};
    \draw[black, fill] (0, 0) circle (0.5pt) node[below] {0};
    \draw[orange, fill] (0.55, 0) circle (0.5pt) node[below] {$\textcolor{orange}{\tau}$};
    \draw[black, fill] (1, 0) circle (0.5pt) node[below] {1};
    
  \end{tikzpicture}
  
  \caption{{
 (Left) Either $\Psi_{jj}(\alpha)$ is \textcolor{MidnightBlue}{strictly increasing} over $\alpha\!\in\!(0,1)$, or it is \textcolor{orange}{not}, with some minimal point $\tau$ of vanishing derivative. We prove the former by showing that no such point $\tau$ exists. (Right) The proof is based on properties of the function~$f(\alpha)$ in \cref{eq:f-convex}. The function is convex; it also satisfies $f(0)\!=\!f(\tau)\! =\! \Psi_{jj}(\tau)$ and $f'(0)f'(\tau)<0$.}
  }
  \label{fig:convex-function}
\end{figure}

We start by rewriting the fixed-point equation in Proposition~\ref{prop:renyi-solution} and eq.~(\ref{eq:fixed-point-var}) in a slightly different form,
\begin{equation}
\Psi_{ii}(\alpha) = [\bphi(\alpha)^{-1}]_{ii} = \left[\alpha\bsigma^{-1} + (1\!-\!\alpha)\Psi^{-1}(\alpha)\right]_{ii}^{-1},
\label{eq:fixed-point-psi}
\end{equation}
where on the right side we have explicitly indicated all sources of dependence on $\alpha$. Next, we differentiate both sides of eq.~(\ref{eq:fixed-point-psi}) for $i\!=\!j$ and at $\alpha\!=\!\tau$:
\begin{equation}
\hspace{-2ex}
\Psi_{jj}'(\tau) = 
       \mathbf{e}_j^\top\!\left\{\frac{d}{d\alpha}\Big[\tau\bsigma^{-1}\!+(1\!-\!\tau)\bpsi^{-1}(\alpha)\Big]^{-1}\!\!
        + \frac{d}{d\alpha}\Big[\alpha\bsigma^{-1}+(1\!-\!\alpha)\bpsi^{-1}(\tau)\Big]^{-1}
       \right\}\bigg|_{\alpha=\tau}\!\!\!\mathbf{e}_j,
\label{eq:dPsiTwoTerms}
\end{equation}
where on the right side we have separated out the different sources of dependence on $\alpha$. Evaluating the first term on the right side, we find that
\begin{equation}
\hspace{-2ex}
 \mathbf{e}_j^\top\!\left\{\frac{d}{d\alpha}\Big[\tau\bsigma^{-1}\!+(1\!-\!\tau)\bpsi^{-1}(\alpha)\Big]^{-1}
       \right\}\bigg|_{\alpha=\tau}\!\!\!\!\mathbf{e}_j =
  \mathbf{e}_j^\top\bphi^{-1}(\tau)\bpsi^{-1}(\tau)\bpsi'(\tau)\bpsi^{-1}(\tau)\bphi^{-1}(\tau)\,\mathbf{e}_j,\hspace{-1ex}
  \label{eq:d-alpha-first-term}
\end{equation}
and now we recognize that this term is nonnegative: of the matrices on the right side of eq.~(\ref{eq:d-alpha-first-term}), note that $\bphi(\alpha)$ and $\bpsi(\alpha)$ are positive definite 
and  $\bpsi'(\tau)$ is positive semidefinite by virtue of the defining properties of $\tau$ in eqs.~(\ref{eq:tau-psd}--\ref{eq:dPsijjtau}).
Dropping this first term from the right side of eq.~(\ref{eq:dPsiTwoTerms}), we obtain the inequality
\begin{equation}
0 \geq  \frac{d}{d\alpha}\left\{
       \mathbf{e}_j^\top\left[\alpha\bsigma^{-1}+(1\!-\!\alpha)\bpsi^{-1}(\tau)\right]^{-1}
       \mathbf{e}_j\right\}\Big|_{\alpha=\tau},
\label{eq:d-alpha-ineq}       
\end{equation}
where we have exploited that the left side of eq.~(\ref{eq:dPsiTwoTerms}) vanishes by the second defining property of $\tau$ in eq.~(\ref{eq:dPsijjtau}). 

\begin{figure}
    \centering
    \includegraphics[width=0.75\linewidth]{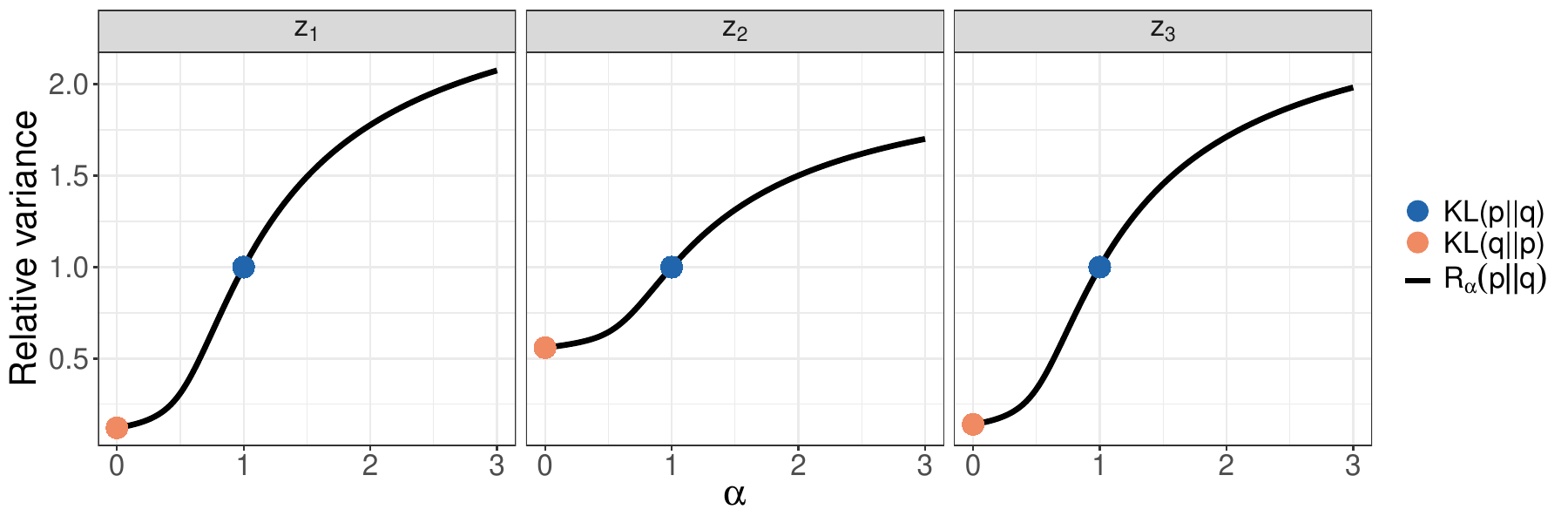}
    \caption{{Marginal variances of $q$ in eq.~(\ref{eq:q_normal}) that minimize $\Ra(q||p)$ as a function of~$\alpha$. The target $p$ is a three dimensional Gaussian. The plot shows the variances of~$q$ normalized by the variances of~$p$. The variances of $q$ are a strictly increasing function of $\alpha$, indicating that the $\alpha$-divergences are ordered.
    While we prove this to be the case for any Gaussian target when $\alpha\! \in\! (0, 1)$, it remains an open problem to prove this when $\alpha\! >\! 1$.}}
    \label{fig:alpha-ordering}
\end{figure}

Next we show that eq.~(\ref{eq:d-alpha-ineq}) contains a contradiction. To do so, we examine the properties of the function defined on $[0,1]$ by
\begin{equation}
  f(\alpha) = \mathbf{e}_j^\top\left[\alpha\bsigma^{-1}+(1\!-\!\alpha)\bpsi^{-1}(\tau)\right]^{-1}
       \mathbf{e}_j.
\label{eq:f-convex}
\end{equation}
This is, of course, exactly the function that is differentiated in eq.~(\ref{eq:d-alpha-ineq}), whose right side is equal to $f'(\tau)$. This function possesses several key properties. First, we note that
\begin{equation}
    f(0) = \mathbf{e}_j^\top\left[\bpsi^{-1}(\tau)\right]^{-1}\mathbf{e}_j = \Psi_{jj}(\tau) = \mathbf{e}_j^\top\left[\tau\bsigma^{-1}+(1\!-\!\tau)\bpsi^{-1}(\tau)\right]^{-1}\mathbf{e}_j = f(\tau).
\end{equation}
Second, we note that $f$ is convex on the unit interval; here, we are observing the convexity of the matrix inverse~\citep{Nordstrom:2011}.
Third, we note that
\begin{equation}
  f'(0) = -\mathbf{e}_j^\top \bpsi(\tau)\left[\bsigma^{-1}\!-\!\bpsi^{-1}(\tau)\right]\bpsi(\tau)\,\mathbf{e}_j 
    = \Psi_{jj}(\tau)\left[\bsigma^{-1}\!-\!\bpsi^{-1}(\tau)\right]_{jj} < 0.
\end{equation}
Finally, we note that the sign of $f'(\tau)$ must \textit{oppose} the sign of $f'(0)$; as shown in the right panel of Figure~(\ref{fig:convex-function}), this follows from the preceding properties that $f(0)\!=\!f(\tau)$, that the function $f$ is convex on $[0,1]$, and that $f'(0)\!<\!0$. Thus we have shown that $f'(\tau)\!>\!0$. But this is directly in \underline{contradiction}  with eq.~(\ref{eq:d-alpha-ineq}), which states that $0\geq f'(\tau)$, and we are forced to conclude that no such point $\tau$ exists. This completes the proof.\mbox{\hspace{23ex}$\blacksquare$}

Combining the results from Sections~\ref{sec:ordering-score}-\ref{sec:ordering-alpha}, we obtain a proof of Theorem~\ref{thm:ordering}.

\begin{mdframed}[hidealllines = true, backgroundcolor = Cerulean!10]
\begin{remark}
The proof in this section does not generalize to $\alpha$-divergences with $\alpha\!>\!1$ because in this case the function $f(\alpha)$ in eq.~(\ref{eq:f-convex}) is not manifestly convex.
\end{remark}
\end{mdframed}

We conjecture that the $\alpha$-divergences are also ordered for $\alpha\! >\! 1$, and we provide empirical evidence for this ordering in Figure~\ref{fig:alpha-ordering} where the target is a three-dimensional Gaussian. 
We leave a proof (or disproof) of this conjecture to future work.

\subsection{Entropy-matching Solution for FG-VI}

  \begin{figure}
    \begin{center}
    \includegraphics[width=0.8\textwidth]{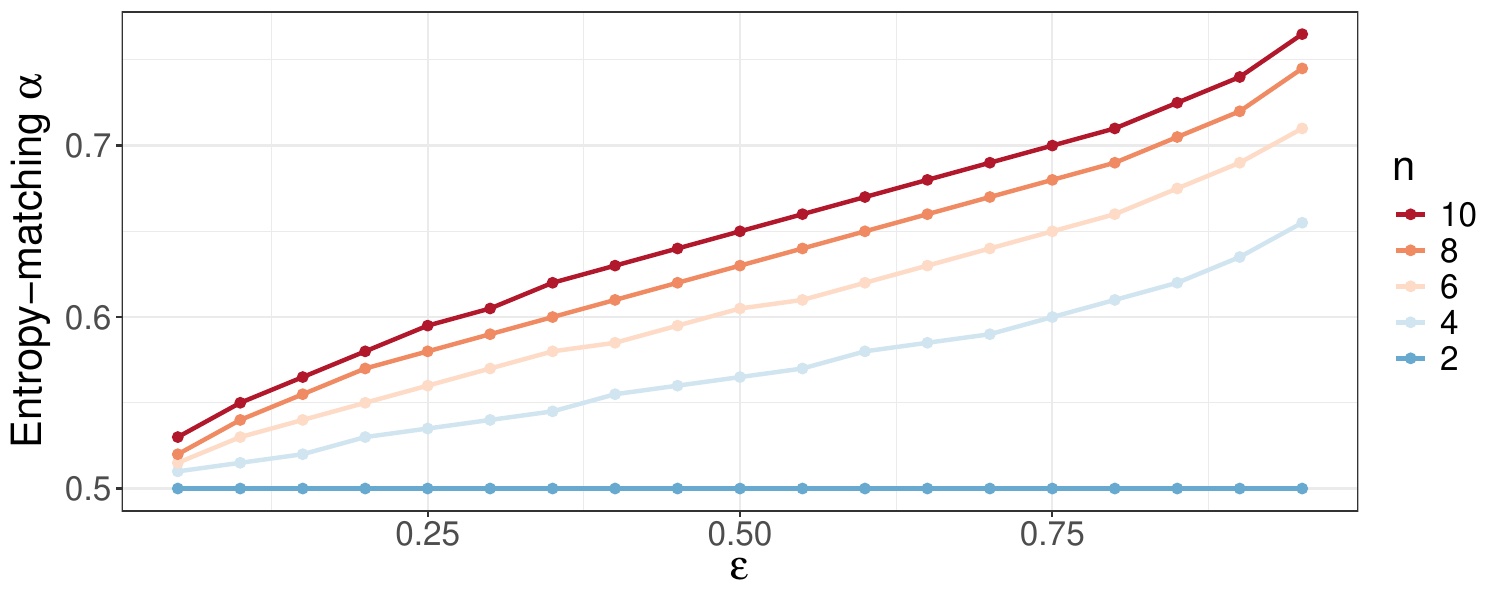}
    \end{center}
    \caption{{When $p$ is Gaussian over $\mathbb{R}^n$, there always exists a unique $\alpha \in (0, 1)$ such that the factorized approximation $q$ minimizing $\Ra(p||q)$ matches the entropy of $p$. The plots shows, however, that the entropy-matching value of $\alpha$ depends on the dimension and covariance structure of $p$. Above we vary the dimension~$n$ and constant correlation $\varepsilon$.}}
    \label{fig:entropy-matching}
  \end{figure}

We observed in \Cref{sec:KL} that the reverse KL divergence yields a precision-matching solution for FG-VI and that the forward KL divergence yields a variance-matching solution. The ordering of divergences in \Cref{thm:ordering} has another interesting consequence: it implies that some $\alpha$-divergence yields a unique \textit{entropy-matching} solution for FG-VI.
  
\begin{mdframed}[hidealllines = true, backgroundcolor = Cerulean!10]
  \begin{corollary}  \label{cor:entropy-matching}
  Let $q_\alpha = \arg\!\min_{q\in\mathcal{Q}} \Ra(p||q)$. 
    There exists a unique value $\alpha \in (0, 1)$ such that $q_\alpha$ matches the entropy of $p$, or equivalently, that $|\bpsi(\alpha)| = |\bsigma|$.
  \end{corollary}
  \end{mdframed}
  \begin{proof}
    Let $\bpsi(\alpha)$ denote the covariance of $q_\alpha$ for $\alpha\!\in\!(0,1)$.
    Since $\Psi_{ii}(\alpha)$ is continuous with respect to~$\alpha$, so is $\log |\bpsi(\alpha)| = \sum_i \log\Psi_{ii}(\alpha)$.
    Moreover, from the ordering of $\alpha$-divergences, 
    we see that $\log|\Psi(\alpha)|$ is not only continuous, but strictly increasing over the unit interval $\alpha\!\in\!(0,1)$. In addition, for all components $i$ along the diagonal, we have
    \begin{equation}
      1 / \Sigma^{-1}_{ii} = \Psi_{ii}(0) \le \Psi_{ii}(\alpha) \le \Psi_{ii}(1) = \Sigma_{ii}, 
    \end{equation}
    with $q_\alpha$ interpolating smoothly between the precision-matching and the variance-matching approximations of FG-VI.
    From the impossibility result in Theorem~\ref{thm:impossibility}, the precision-matching approximation \mbox{(at $\alpha\!=\!0$)} underestimates the entropy of $p$, while the variance-matching approximation \mbox{(at $\alpha\!=\!1$)} overestimates it. From continuity and strict monotonicity, it follows that there exists a unique $\alpha\!\in\!(0,1)$ such that $|\bpsi(\alpha)| = |\bsigma|$, or equivalently, such that $q_\alpha$ matches the entropy of $p$.
  \end{proof}

  \begin{figure}
    \includegraphics[width=0.95\linewidth]{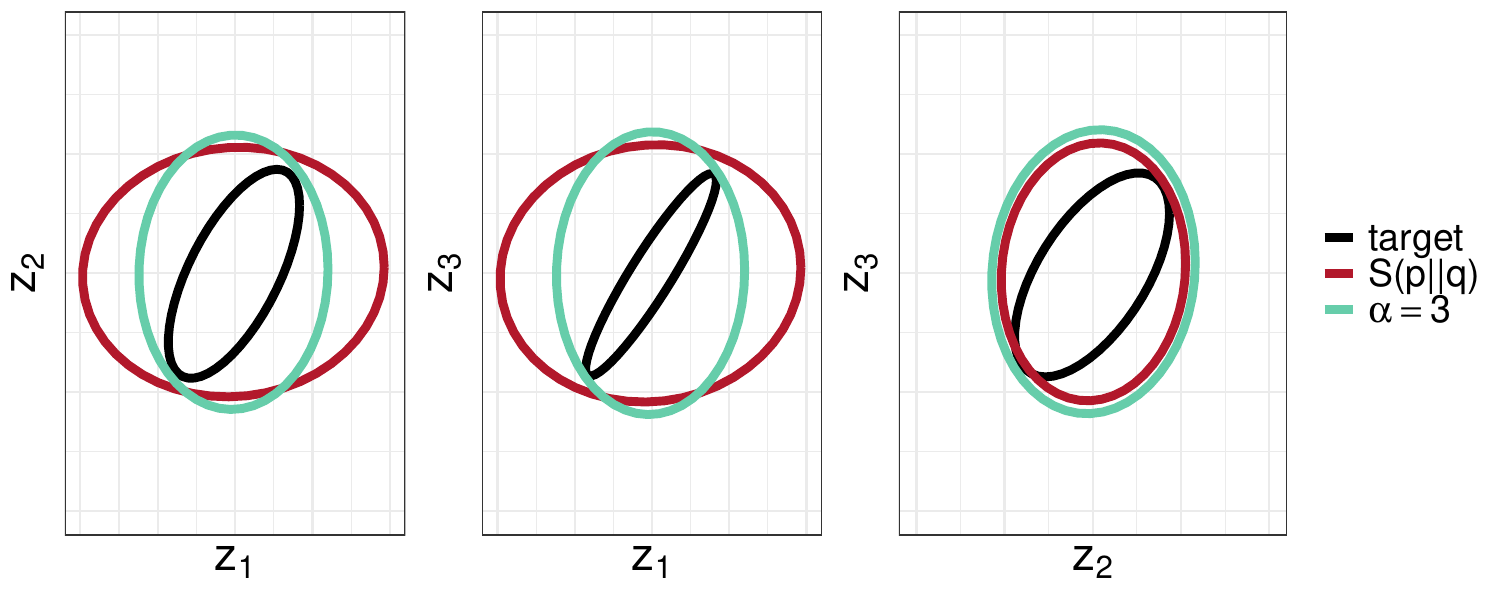}
    \caption{{Example of a Gaussian target in dimension $d=3$ where the variances obtained by minimizing $S(p||q)$ and $\Ra(p||q)$ are \underline{not} ordered.
    The non-ordering holds for any $\alpha \gtrapprox 2$ and is most clearly seen for $\alpha=3$.
    The figure shows the two-dimensional projections of the variational fits.    
    }}
    \label{fig:renyi-solution}
\end{figure}
  
  The proof of Corollary~\ref{cor:entropy-matching} demonstrates the existence an entropy-matching solution, but it is not a constructive proof. Unfortunately, 
  there does not exist a single value of~$\alpha$ for which $|\bpsi(\alpha)| = |\bsigma|$, even when $p$ is Gaussian. We show this numerically for the case where the target covariance $\bsigma$ has unit diagonal elements and constant off-diagonal elements $\varepsilon\!\in\!(0,1)$.   As seen in Figure~\ref{fig:entropy-matching}, the entropy-matching value of $\alpha$ (obtained by a grid search) changes as we vary the amount of correlation and the dimension of the problem.

\subsection{Non-ordering of the Score-based and $\alpha$-divergences for $\alpha\! >\! 1$}

Given that both $S(p||q)$ and $\Ra(p||q)$ dominate $\KL(p||q)$, it is natural to ask whether an ordering exists between the score-based and $\alpha$-divergences when $\alpha\! >\! 1$.
Here we produce a counterexample to show that no such ordering exists.
Searching through randomly generated covariance matrices, we find an example in dimension $d\!=\!3$ where the variances that minimize $\Ra(p||q)$ and $S(p||q)$ are not ordered for $\alpha\!\in\![2,3]$; Figure~\ref{fig:renyi-solution} illustrates this non-ordering when $\alpha\!=\!3$. The target covariance in this example is (within two digits)
\begin{equation}  \label{eq:Gaussian-cov}
    \bsigma = \begin{pmatrix}
        1 & 0.66 & 0.93 \\
        0.66 & 1 & 0.59 \\
        0.93 & 0.59 & 1
    \end{pmatrix}.
\end{equation}
Table~\ref{tab:variances} reports the variances of $q$ obtained by minimizing $S(p||q)$ and $\text{D}_\alpha(p||q)$ for this example when $\alpha\!=\!2$ and $\alpha\!=\!3$.
Note that the score-based divergence yields the largest estimate (shown in bold) of the variance $\Psi_{11}$, but the $\alpha$-divergence (with $\alpha\!=\!3$) yields the largest estimate of the variances $\Psi_{22}$ and~$\Psi_{33}$. Thus these divergences are not ordered.

\begin{table}[]
    \centering
    \begin{tabular}{r r r r}
       \rowcolor{Cerulean!10} {\bf Divergence}  & \hspace{10ex}$\Psi_{11}$ & \hspace{10ex}$\Psi_{22}$ & 
       \hspace{10ex}$\Psi_{33}$  \\
       $S(p||q)$  & {\bf 6.18} & 1.40 & 1.63 \\
       \rowcolor{Cerulean!10} $\text{D}_{\alpha=2}(p||q)$ & 1.78 & 1.50 & 1.71 \\
       $\text{D}_{\alpha=3} (p||q)$ & 2.07 & {\bf 1.70} & {\bf 1.98}
    \end{tabular}
    \caption{{Variances estimated by minimizing different divergences for the Gaussian target in eq.~(\ref{eq:Gaussian-cov}). In bold, the largest optimal variance obtained across all divergences. This example shows that $S(p||q)$ and $\Ra(p||q)$ cannot be ordered for~\mbox{$\alpha\!>\!1$}.}}
    \label{tab:variances}
\end{table}


\section{Numerical Experiments} \label{sec:experiments}

Does the ordering of divergences hold when FG-VI is applied to non-Gaussian targets?
We study this question empirically on a range of target distributions.
Several of these models are taken from Bayesian analysis problems, where the target $p$ is only known up to a normalizing constant.

\subsection{Optimization Procedure}

When $p$ is not Gaussian, some divergences in Table~\ref{tab:divergences} are harder to minimize numerically than others.
We use the following approaches. \\

\noindent
\fcolorbox{Cerulean!10}{Cerulean!10}{\textit{Optimization of $\KL(q||p)$}.} We minimize $\KL(q||p)$ by equivalently maximizing the evidence lower bound (ELBO), and we estimate the ELBO via Monte Carlo with draws from~$q$, a procedure at the core of ``black-box'' VI~\citep[e.g.][]{Kucukelbir:2017}. \\

\noindent
\fcolorbox{Cerulean!10}{Cerulean!10}{\textit{Optimization of $\Ra(q||p)$}.} We use a similar procedure to optimize the $\alpha$-divergence with $\alpha \in \{0.1, 0.5, 2\}$; however, this procedure is more fraught\footnote{For more discussion on implementing VI with $\alpha$-divergences, see e.g., \citet{Hernandez-Lobato:2016, Li:2016, Dieng:2017, Daudel:2021, Daudel:2023b}.} as Monte Carlo estimators of the $\alpha$-divergence and its gradients suffer from a large variance, especially as the dimension increases and as $\alpha$ approaches~1 \citep{Geffner:2021}.
For $\alpha\!=\! 2$, we find it helpful to construct Monte Carlo estimators with \textit{oracle samples} from $p$, rather than $q$, when such samples are available (e.g., for simpler targets). \\

\noindent
\fcolorbox{Cerulean!10}{Cerulean!10}{\textit{Optimization of $\KL(p||q)$}.}
We minimize $\KL(p||q)$ by choosing $q$ in eq.~(\ref{eq:q_normal}) to match the mean and marginal variances of~$p$. When possible, we calculate these statistics of $p$ analytically; otherwise we estimate them by long runs of Markov chain Monte Carlo. \\

\noindent
\fcolorbox{Cerulean!10}{Cerulean!10}{\textit{Optimization of $\SM(q||p)$}.} We minimize $\SM(q||p)$ by adapting a recent \textit{batch-and-match} (BaM) method inspired by proximal point algorithms~\citep{Cai:2024}. BaM was originally developed for Gaussian variational families with dense covariance matrices, but here we implement BaM updates for FG-VI (derived in Appendix~\ref{app:fbam}).
We do not empirically evaluate the forward score-based divergence as we do not have a reliable method to minimize $\SM(p||q)$. \\

Each method is implemented in \texttt{Python} and uses the \texttt{JAX} library to calculate derivatives \citep{jax2018github}.
Optimization is performed with the \texttt{Adam} optimizer \citep{Kingma:2015}.
As a baseline, we run the optimizer for 500 iterations, and at each iteration, we use $B = 10,000$ draws from $q$ (or $p$ when using oracle samples) to estimate the relevant objective function and its gradient.
This large number of draws is overkill for many problems, but it helps stabilize the optimization of $\alpha$-divergences.
In addition, we parallelize many calculations on GPU, keeping run times short.

For certain targets, we find it necessary to fine-tune the optimizer in order to avoid numerical instability.
We do so by adjusting the learning rate of \texttt{Adam}, the number $B$ of Monte Carlo draws, and the number of iterations.
This proves particularly important when minimizing the $\alpha$-divergences.
Even so, we are only able to optimize the $\alpha$-divergences for low-dimensional targets ($D\!\le\!10$).
 Code to reproduce the experiments can be found at \url{https://github.com/charlesm93/VI-ordering}.  

\begin{figure}
    \centering
    \includegraphics[width=0.75\linewidth]{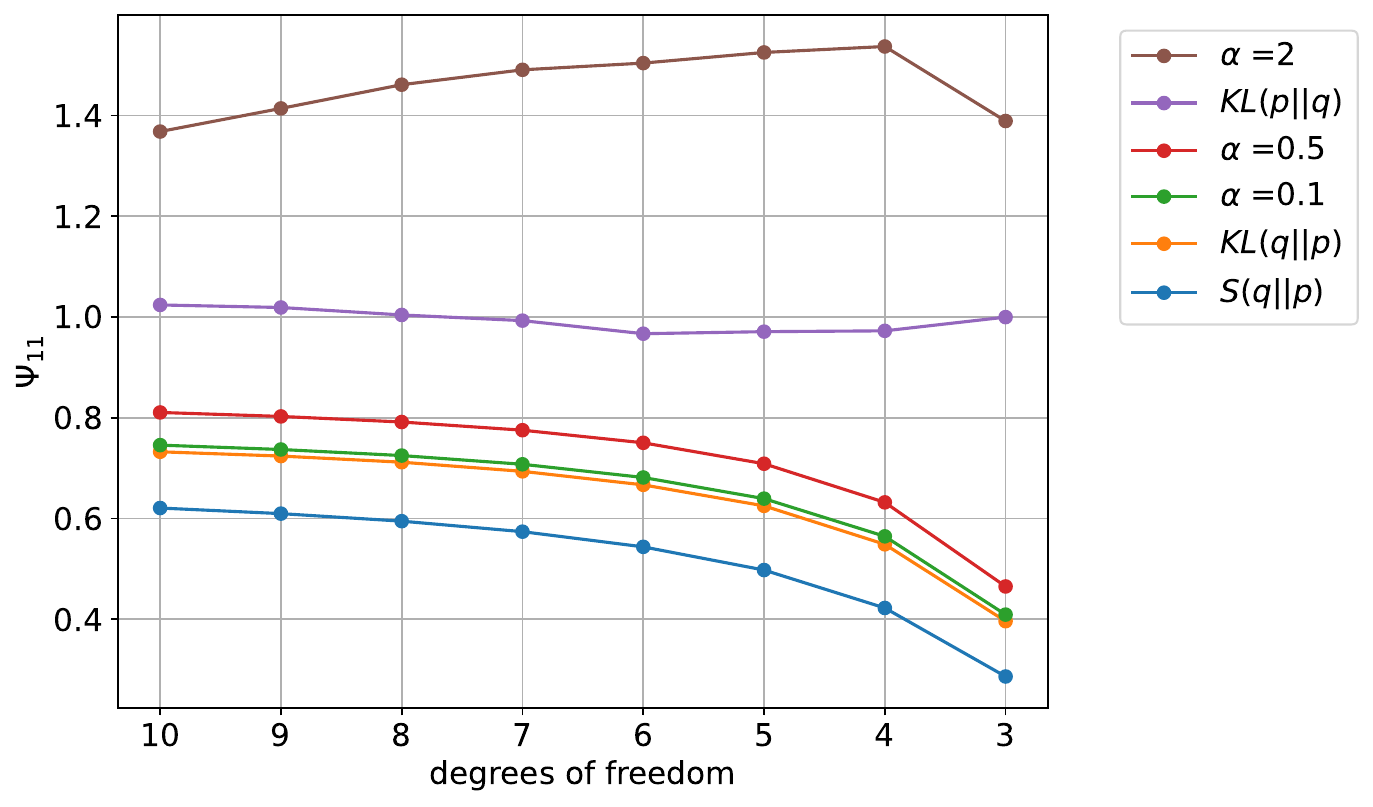}
    \caption{{Variance estimated by FG-VI when targeting a two-dimensional Student-t with varying degrees of freedom~$d_f$. The  ordering of divergences for Gaussian targets is observed across a wide range of Student-t targets, from $d_f\!=\!10$, where the target is nearly Gaussian, to $d_f\!=\!3$, where the target is heavy tailed. (A similar plot is obtained for $\Psi_{22}$.)}}
    \label{fig:student-t}
\end{figure}

\subsection{Targets with Heavy Tails and Skew}

First we experiment with non-Gaussian targets in two dimensions where we can systematically vary the degree of non-Gaussianity. 

The Student-t distribution has heavier tails than the Gaussian, particularly for low degrees of freedom ($d_f$).
We experiment with FG-VI on a wide range of Student-t targets, with $d_f$ ranging from 10 (near Gaussian) to 3 (heavy-tailed), and we find that the ordering of variances in Theorem~\ref{thm:ordering}, derived for Gaussian targets, is preserved when the target is a Student-t distribution; see Figure~\ref{fig:student-t}.

\begin{figure}
    \centering
    \includegraphics[width=0.75\linewidth]{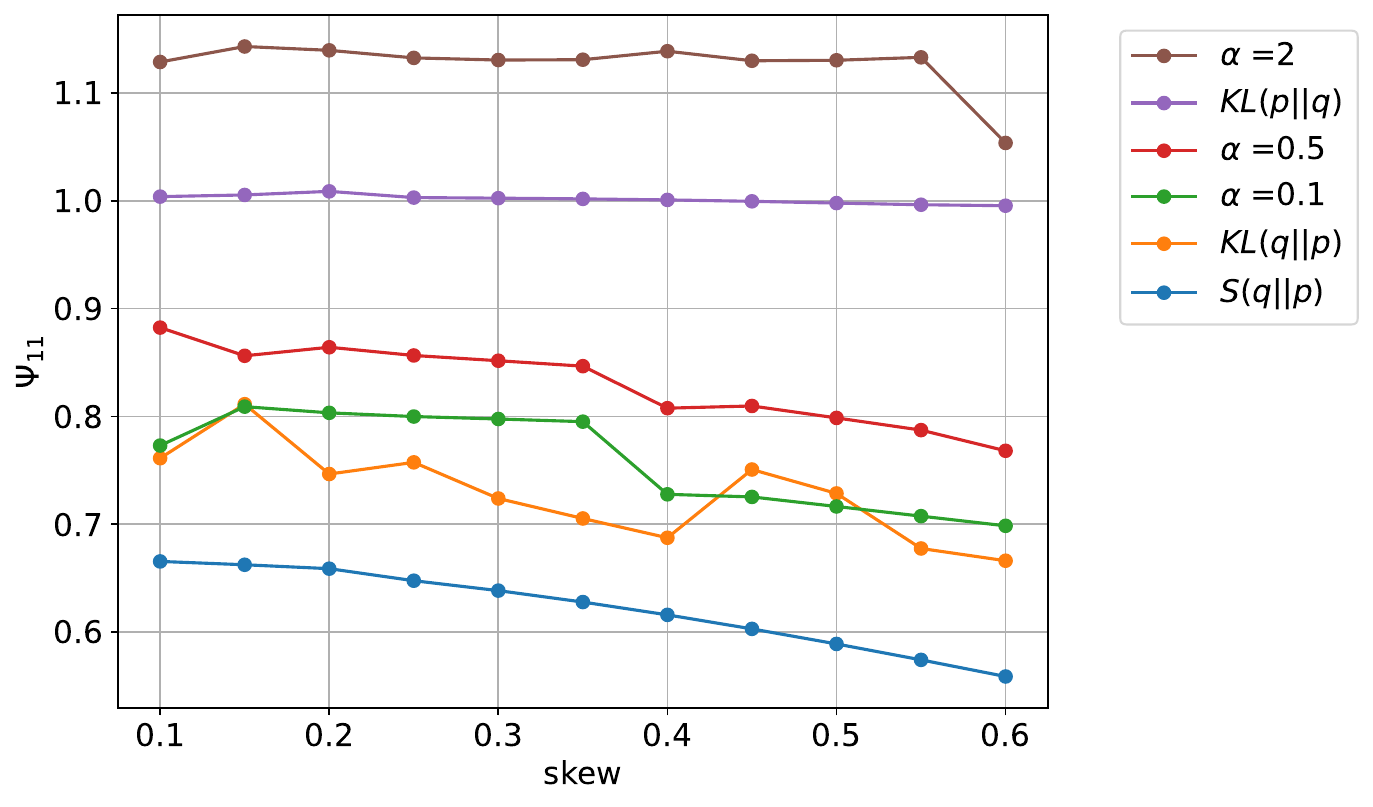}
    \caption{{Variance of FG-VI when targeting a two-dimensional skewed Normal with skewness $s$.
    For $s=0.1$, the target is near Gaussian and $s$ increases, it becomes more asymmetric.
    (A similar plot is obtained for $\Psi_{22}$.)}}
    \label{fig:skew-Normal}
\end{figure}

The multivariate Skew-Normal distribution \citep{Azzalini:1996} has a skewness parameter~$s$: for $s\!=\!0$, the Skew-Normal exhibits no skew and reduces to a Gaussian, whereas for $s \!\ge\! 0.5$ it is heavily asymmetric.
For skewed targets in this family, we find that the ordering of variances in Theorem~\ref{thm:ordering} is generally preserved even as $s$ increases; see Figure~\ref{fig:skew-Normal}.
However, the ordering between $\KL(q||p)$ and $\Ra(q||p)$ for $\alpha = 0.1$ is sometimes violated. These violations are slight and may be due to numerical error in the optimization of these divergences.

\begin{table}
    \begin{center}
        \begin{tabular}{l r l c c}
        \rowcolor{Cerulean!10} {\bf Model} & $n$ & {\bf Details} & {\bf Variances} & {\bf Entropy}  \\
        Gaussian & 10 & Full-rank covariance matrix. & \cmark & \cmark \\
        \rowcolor{Cerulean!10} Student-t & 2 & Heavy tail. & \cmark & \cmark \\
        Skew-Normal & 2 & Asymmetric. & \textbf{?} & \textbf{?} \\ 
        \rowcolor{Cerulean!10} Rosenbrock & 2 &  High curvature. & \cmark & \cmark \\
        Eight Schools & 10 & Small hierarchical model with & \xmark & \cmark \\
        & & funnel geometry. \\
        \rowcolor{Cerulean!10} German Credit & 25 & Logistic regression. & \cmark & \cmark \\
        Radon Effect & 91 & Large hierarchical model with & \xmark & \cmark \\
        & & funnel geometry. \\
        \rowcolor{Cerulean!10} Stochastic Volatility & 103 &  Stochastic volatility model. & \xmark & \cmark
        \end{tabular}
        \caption{
        {Target distributions $p$ with dimension $n$. The two right-most columns indicate respectively whether the divergences are ordered according to the marginal variances (i.e., the ordering in Theorem~\ref{thm:ordering} is preserved) or according to the entropy (equivalently, the generalized variance). Question marks indicate that the ordering in Theorem~\ref{thm:ordering} is only slightly violated, possibly due to noise in the stochastic~optimization.
        }}
        \label{tab:targets}
    \end{center}
\end{table}

\subsection{Models from the \texttt{Inference Gym}}

We next experiment with 
target distributions from the \texttt{inference gym}, a curated library of diverse models for the study of
inference algorithms \citep{inferencegym2020}.
We describe these target distributions briefly in Table~\ref{tab:targets} and provide more details in Appendix~\ref{app:models}.

\begin{figure}[t]
    \begin{center}
    \includegraphics[width=5in]{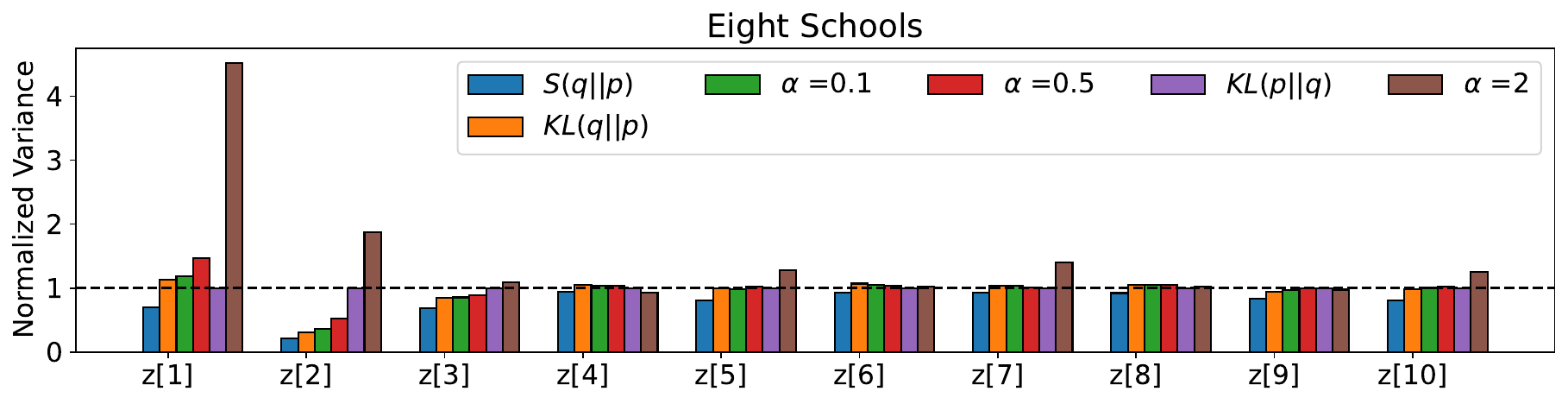}
    \end{center}
    \caption{
    {Variances from FG-VI when targeting the posterior distribution of the Eight Schools model.
    Some of the variances are ordered in the same way as predicted for Gaussian targets, but most are not.
    In this case, the target posterior is highly non-Gaussian, exhibiting a funnel shape, and thus strongly violating the assumptions of Theorem~\ref{thm:ordering}. The variances reported here, unlike those appearing in Theorem~\ref{thm:ordering}, are also noisy estimates from a stochastic optimization.
    }
    }
    \label{fig:8schools_variance}
\end{figure}

Some of these target distributions are far from Gaussian, and so the ordering of variances in Theorem~\ref{thm:ordering} is not guaranteed to hold. We may also observe slight violations of this ordering due to noisy solutions that arise from stochastic optimization.
%

Violations of the ordering in Theorem~\ref{thm:ordering} are observed for the Eight Schools model, a hierarchical model that exhibits a funnel geometry; see Figure~\ref{fig:8schools_variance}. In this example, most of the marginal variances are not ordered in the same way as predicted for Gaussian targets, although in many cases, the violations are slight.
(The true marginal variances in this model were estimated by long runs of Markov chain Monte Carlo.)

\begin{figure}[t]
    \centering
    \includegraphics[width=1.7in]{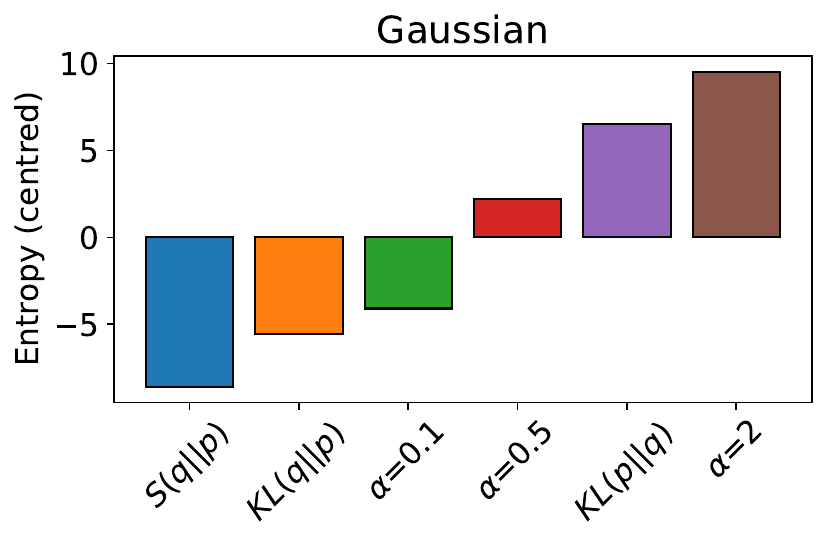}
    \includegraphics[width=1.7in]{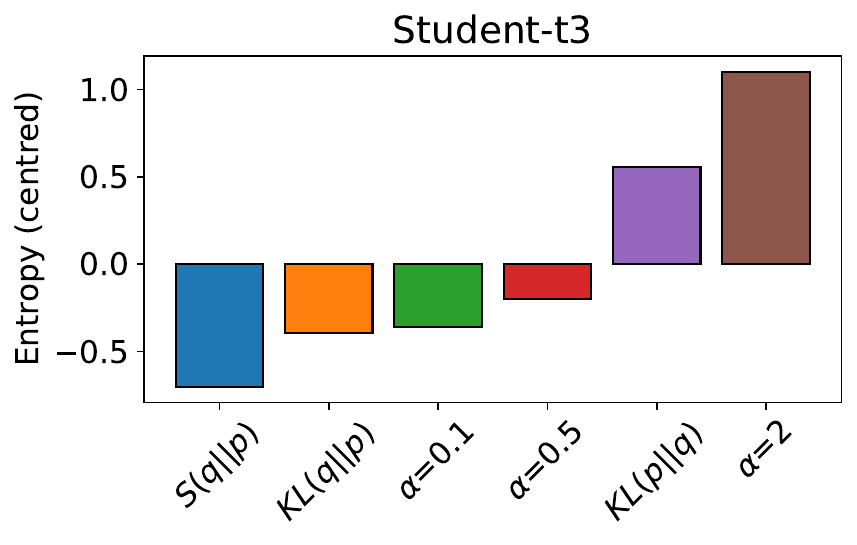}
    \includegraphics[width=1.7in]{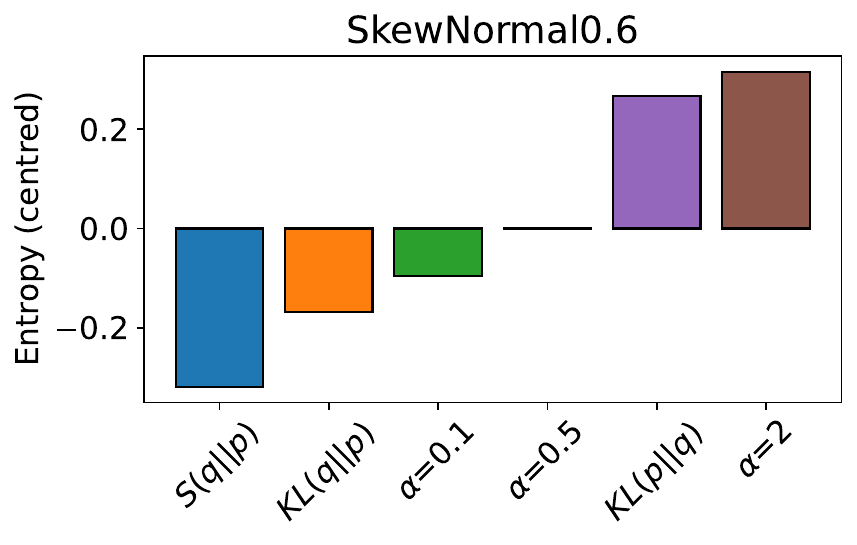}
    \includegraphics[width=1.7in]{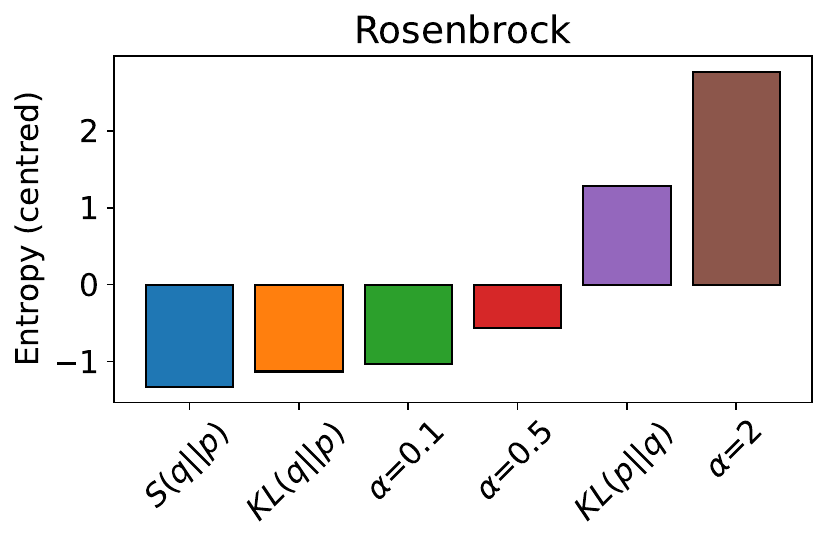}
    \includegraphics[width=1.7in]{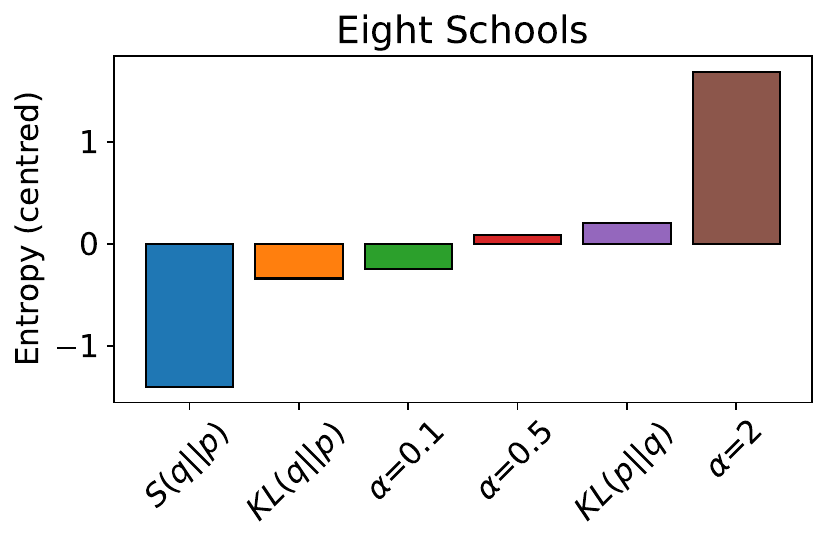} \\
    \includegraphics[width=1.7in]{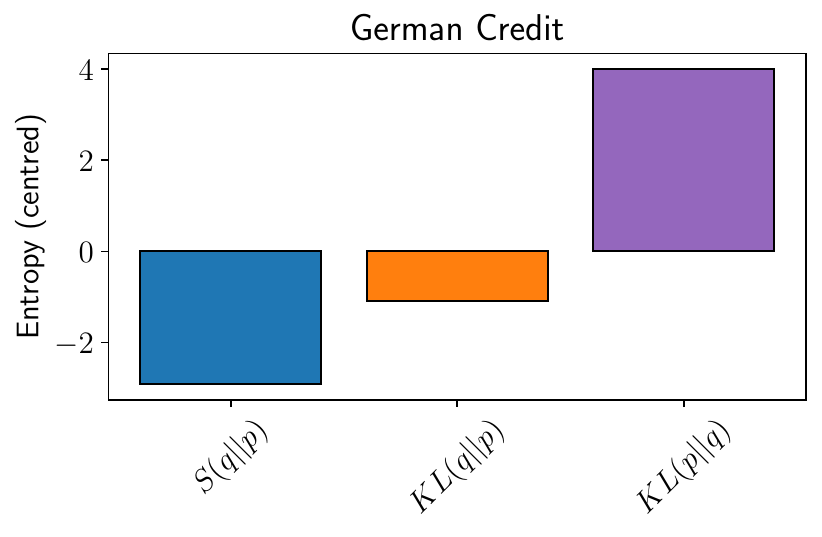}
    \includegraphics[width=1.7in]{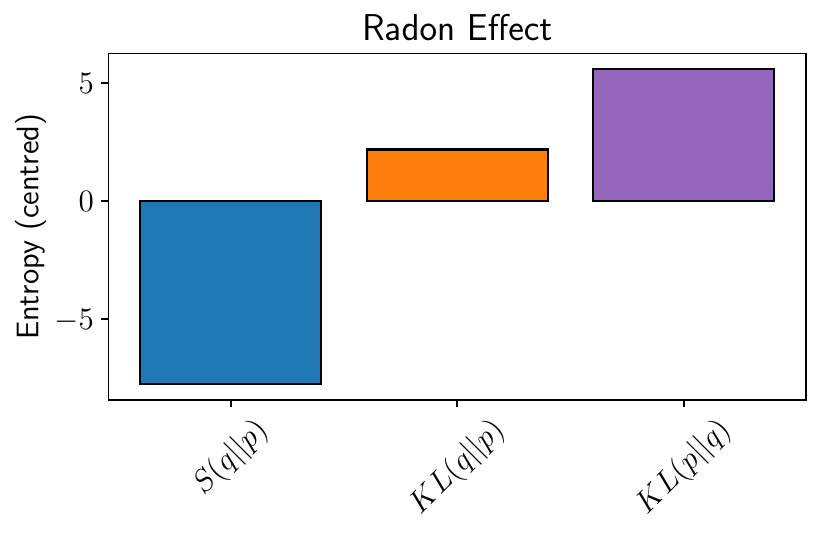}
    \includegraphics[width=1.7in]{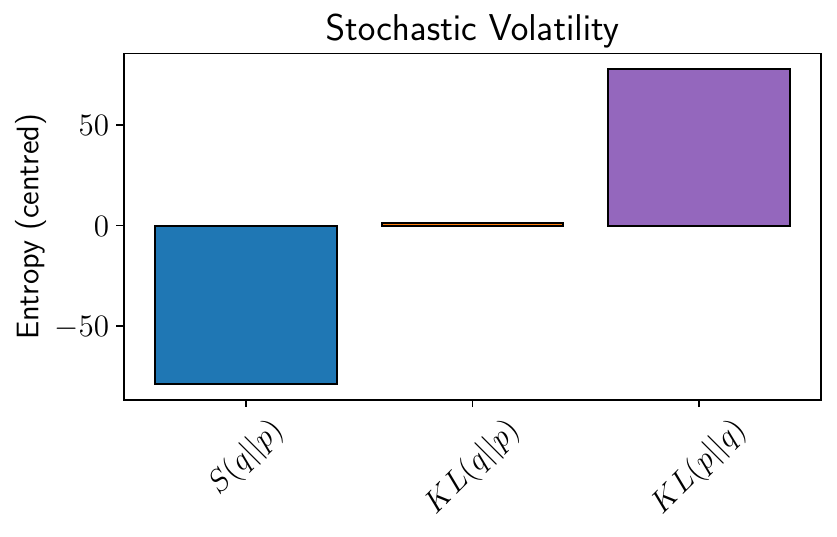}
    
    \caption{
    {
    Ordering of entropy across targets. The zero on each vertical axis corresponds to the average of the entropies estimated by the approximations in each panel.
    }}
    \label{fig:entropy-non-Gaussian}
\end{figure}

When $p$ is not Gaussian, an alternative ordering of divergences for FG-VI is suggested by the estimates they yield of the joint entropy. As shown in the final column of Table~\ref{tab:targets} and in Figure~\ref{fig:entropy-non-Gaussian}, the ordering in \Cref{thm:ordering} \textit{does} correspond---across all of the non-Gaussian targets in our experiments---to the ordering of entropies estimated by FG-VI. 
In these experiments, however, we are not able to compare the entropies estimated by FG-VI to the actual entropies. To compute the entropy of non-Gaussian distributions, it is necessary to estimate their normalizing constants; this can be done by invoking a Gaussian-like approximation, as in bridge sampling \citep{Meng:2002, Gronau:2020}. However, it seems inadequate to compare a theory developed 
for Gaussian targets to 
an empirical benchmark that 
relies on a Gaussian approximation.

\section{Discussion}
\label{sec:discuss}

Divergences over probability spaces are ubiquitous in 
the statistics and machine learning literature, but it remains challenging to reason about them.
This challenge is salient in VI, where the divergence does not serve merely as an object for theoretical study but also bears directly on the algorithmic implementation of the method.
Our work reveals how the choice of divergence can align (or misalign) with intelligible inferential goals, and it is in line with a rich and recent body of work on assessing the quality of VI \citep[e.g,][]{Yao:2018, Wang:2018, Huggins:2020, Dhaka:2021, Biswas:2024}.
A distinguishing feature of our paper is that it analyzes multiple and competing inferential goals---goals which cannot be simultaneously met and which are each best served by a different divergence.

Our results can inform the choice of divergence to minimize in eq.~(\ref{eq:argminDqp}) in tandem with the use of a factorized approximation. This choice will in turn depend on the problem at hand. We discuss some examples:
\begin{itemize}
    

    \item {\bf Bayesian statistics.} Here, uncertainty is primarily quantified by the marginal variances~$\Sigma_{ii}$, and these are matched by minimizing $\KL(p||q)$. In general, however, this divergence is not straightforward to minimize, though strategies such as expectation-propagation can in some limits attain an optimal solution \citep{Vehtari:2020}. It is much more straightforward to minimize divergences such as $\KL(q||p)$ and $S(q||p)$, but for FG-VI we have seen how marginal variances can be significantly underestimated by factorized approximations that minimize these divergences. This provides further motivation for: (i) corrective procedures~\citep{Giordano:2018} to adjust the estimate of marginal variances from F-VI; (ii) VI methods with richer (non-factorized) families \citep[e.g][]{Opper:2009, Dhaka:2021, Zhang:2022}; (iii) altogether alternative methods such as Markov chain Monte Carlo \citep[MCMC;][]{Robert:2004}, when allowed by a computational budget.  
    
    \item {\bf Pre-conditioning of Markov chain Monte Carlo.} Several MCMC algorithms benefit from transforming the latent variables over which the Markov chain is constructed. This technique is known as pre-conditioning.
    In Hamiltonian Monte Carlo \citep[HMC; ][]{Neal:2012, Betancourt:2018} the mass matrix serves as a pre-conditioner, and one typically restricts it to be diagonal for computational reasons.
    \citet{Neal:2012} recommends setting the diagonal elements of the (inverted) mass matrix to early estimates of the marginal variances, 
    an idea implemented in popular adaptive HMC algorithms \citep[e.g][]{Hoffman:2014}.
    Recently, it was proposed that this approach might not be optimal \citep{Hird:2023}, 
    and that a better choice for a diagonal mass matrix might be based on estimates of the marginal precisions \citep{Tran:2024}.
    This suggests a novel application of F-VI, one where it is used to tune HMC's mass matrix by minimizing $\KL(q||p)$.
    Investigation of this method is left as future work.

    \item {\bf Information theory.} Here uncertainty is primarily measured by the entropy. 
    In certain limiting cases, the entropy may be accurately estimated by minimizing $\KL(q||p)$ 
    \citep{Parisi:1988, Margossian:2023}; in general, however, this is not the case, as we have shown analytically when $p$ and $q$ are Gaussian.
    One possible strategy in this setting is to instead minimize the $\alpha$-divergence, for $\alpha \in (0, 1)$.
    While there always exists an $\alpha$ to match the entropy, we have shown that this value depends on $p$, and so it remains an open problem to find 
    an actionable divergence for entropy estimation. 
\end{itemize}

VI is also deployed in applications where the primary goal is not to quantify uncertainty.
A prominent example is latent variable models, where VI is used to approximate a marginal likelihood.
In this context, the KL and $\alpha$-divergences have been ordered based on the (lower or upper) bound they provide on the marginal likelihood \citep{Li:2016, Dieng:2017}. (Interestingly, this ordering agrees with the one provided in this paper, even though it is based on a different criterion.) This type of ordering cannot be extended to score-based divergences, however, as they do not provide bounds on the marginal likelihood.

We have seen that the results of VI can be improved, for certain inferential goals, by choosing a particular divergence. 
A complementary and more common approach is to use 
a richer family $\mathcal Q$ of variational approximations \citep[e.g,][]{Hoffman:2015, Johnson:2016, Dhaka:2021}.
The deficits of FG-VI may be addressed, for instance, by Gaussian variational approximations with full covariance matrices~\citep{Opper:2009,Cai:2024}.
In high dimensions, \citet{Dhaka:2021} argue that the results of VI can be improved more easily by optimizing $\KL(q||p)$ over a richer variational family than by optimizing certain alternative divergences.
%
Nevertheless, even as the variational family becomes richer, it is not likely to contain the target distribution~$p$, and therefore the approximating distribution $q$ must in some way be compromised. For example, if $\mathcal Q$ is a family of symmetric distributions and the target $p$ is asymmetric, then an approximation will not be able to match both the mode \textit{and} the mean of the target. It stands to reason that further impossibility theorems, in the spirit of Theorem~\ref{thm:impossibility}, can be derived to elucidate these trade-offs. 

%
\section*{Acknowledgments}


 We thank David Blei, Diana Cai, Robert Gower, Chirag Modi, and Yuling Yao for helpful discussions.
We are grateful to anonymous reviewers from the conference on Uncertainty in Artificial Intelligence for their comments on a previous paper \citep{Margossian:2023}, which provided some of the motivation for this manuscript.
Finally we thank the editor, Barbara Engelhardt, and two anonymous reviewers for their feedback on this manuscript.



\appendix



\section{Batch and Match Algorithm for FG-VI} \label{app:fbam}

\textit{Batch and Match} (BaM) \citep[BaM]{Cai:2024} is an iterative algorithm for VI that attempts to estimate and minimize 
the reverse score-based divergence, $\SM(q||p)$. 
At each iteration, BaM minimizes a regularized objective function
\begin{equation} \label{eq:regularized-obj}
    \mathcal L^{\text{BaM}}(q) = \widehat{\mathcal D}_{q_t}(q||p) + \frac{2}{\lambda_t} \KL(q || q_t),
\end{equation}
where
\begin{itemize}
    \item $q_t \in \mathcal Q$ is a current approximation,
    \item $\hat{\mathcal D}_{q_t}(q||p)$ is a Monte Carlo estimator of the score-based divergence
    \begin{equation}
    \hat{\mathcal D}_{q_t}(q||p) = \frac{1}{B} \sum_{b = 1}^B ||\nabla\log q\big(\bz^{(b)}\big) - \nabla\log p\big(\bz^{(b)}\big)||^2_{\bpsi},
    \end{equation}
    using draws $\bz^{(b)} \sim q_t$.
    \item $\lambda_t > 0$ is the learning rate, or step size, at the $t^\text{th}$ iteration. 
\end{itemize}
\citet[Appendix C]{Cai:2024} derive the BaM updates which minimize \cref{eq:regularized-obj} when $\mathcal Q$ is the family of 
Gaussian distributions with \textit{dense} covariance matrices.
In this appendix, we derive the analogous updates to minimize \cref{eq:regularized-obj} when
$\mathcal Q$ is the family of Gaussian distributions with \textit{diagonal} covariance matrices. These were the updates used for the experiments in \Cref{sec:experiments} of the paper.

We follow closely the derivation in \citet{Cai:2024}, noting only the essential differences for the case of FG-VI.
We use ${\bf g}^{(b)} = \nabla \log p \big( {\bf z}^{(b)}\big)$ as shorthand for the score at the $b^\text{th}$ sample.
As before, the following averages need to be computed at each iteration:
\begin{equation} \label{eq:bam-quantities}
    \bar {\bf z} = \frac{1}{B} \sum_{b=1}^B {\bf z}^{(b)}, \quad
    \bar {\bf g} = \frac{1}{B} \sum_{b=1}^B {\bf g}^{(b)}.
\end{equation}
For FG-VI, we instead compute \textit{diagonal} matrices ${\bf C}$ and $\boldsymbol{\Gamma}$ with nonnegative elements
\begin{equation}
    C_{ii} = \frac{1}{B} \sum_{b=1}^B \left (z_i^{(b)} - \bar z_i \right)^2, \ \ \Gamma_{ii} = \frac{1}{B} \sum_{b=1}^B \left (g^{(b)}_i - \bar g_i \right)^2.
\end{equation}
Let $q_{t + 1}$ denote the minimizer of \cref{eq:regularized-obj}, with mean $\bnu_{t+1}$ and diagonal covariance matrix~$\bpsi^{t + 1}$.
The update for the mean takes the same form as in the original formulation of~BaM:
%
\begin{equation}
    \bnu^{t + 1} = \frac{\lambda_1}{1 + \lambda_t} \left (\bar {\bf z} + \bpsi^{t + 1} \bar {\bf g} \right) + \frac{1}{1 + \lambda_t} \bnu^t. 
\end{equation}
To obtain the covariance update, we substitute this result into \cref{eq:regularized-obj} and differentiate with respect to the \textit{diagonal} elements of $\bpsi^{t+1}$. For FG-VI, we obtain a simple quadratic equation for each updated variance:
%
\begin{equation} \label{eq:bam-quadratic}
    \left (\Gamma_{ii} + \frac{1}{1 + \lambda_t} \bar g_i^2 \right) \left ( \Psi_{ii}^{t + 1} \right)^2 + \frac{1}{\lambda_t} \Psi^{t + 1}_{ii} - \left (C_{ii} + \frac{1}{\lambda_t} \Psi^t_{ii} + \frac{(\nu^t_i - \bar z_i)^2}{1 + \lambda_t} \right) = 0. 
\end{equation}
Eq.~(\ref{eq:bam-quadratic}) always admits a positive root, and for FG-VI, this positive root is the BaM update for the $i^\text{th}$ diagonal element of $\bpsi^{t+1}$.

At a high level, BaM works the same way with diagonal covariance matrices as it does with full covariance matrices.
Each iteration involves a \textit{batch} step which draws the $B$ samples from $q_t$ and a \textit{match} step that performs the calculations in eqs.~(\ref{eq:bam-quantities}-\ref{eq:bam-quadratic}).
We find a constant learning rate $\lambda_t\! =\! 1$ to work well for the experiments in \Cref{sec:experiments}.

\section{Models from the \texttt{Inference Gym}} \label{app:models}

In this appendix, we provide details about the models and data sets from the \texttt{inference gym} used in \Cref{sec:experiments}. \\

\noindent
\textit{Rosenbrock Distribution.} (n\! =\! 2) A transformation of a normal distribution, with non-trivial correlations \citep{Rosenbrock:1960}. 
The contour plots of the density function have the shape of a crescent moon.
The joint distribution is given by:
\begin{align}
    p(z_1) & = \text{normal} (0, 10) \\
    p(z_2 | z_1) & = \text{normal}(0.03 (z_1^2 - 100), 1).
\end{align}
We use FG-VI to approximate $p(z_1,z_2)$. \\

\noindent
\textit{Eight Schools.} ($n\!=\!10$)
A Bayesian hierarchical model of the effects of a test preparation program across 8 schools \citep{Rubin:1981}:
\begin{align}
    p(\mu) & = \text{normal}(5, 3^2) \\
    p(\tau) & = \text{normal}^+(0, 10) \\
    p(\theta_i|\mu, \sigma) & = \text{normal}(\mu, \tau^2) \\
    p(y_i| \theta_i) & = \text{normal}(\theta_i, \sigma_i^2),
\end{align}
where $\text{normal}^+$ is a normal distribution truncated at 0.
The observations are ${\bf x} = (y_{1:8}, \sigma_{1:8})$ and the latent variables are ${\bf z} = (\mu, \tau, \theta_{1:8})$.
We use VI to approximate the posterior distribution $p(\mathbf{z}|\mathbf{x})$. \\

\noindent
\textit{German Credit.} ($n\! =\! 25$)
A logistic regression applied to a credit data set \citep{Dua:2017}, with covariates ${\bf x} = x_{1:24}$, observations ${\bf y} = y_{1:24}$, coefficients $\bz = z_{1:24}$ and an intercept $z_0$:
\begin{align}
    p(\bz) & = \text{normal}(0, {\bf I}_n), \\
    p(y_i | \bz, {\bf x}) & = \text{Bernoulli} \left ( \frac{1}{e^{- \bz^T {\bf x}_i - z_0}}  \right)
\end{align}
We use FG-VI to approximate $p(\mathbf{z}|y_i,\mathbf{x})$. \\

\noindent
\textit{Radon Effect.} ($n\!=\!91$)
A hierarchical model to measure the Radon level in households \citep{Gelman:2007}.
Here, we restrict ourselves to data from Minnesota.
For each household, we have three covariates, ${\bf x}_i = x_{i,1:3}$, with corresponding coefficients ${\bf w} = w_{1:3}$, and a county level covariate, $\theta_{j[i]}$, where $j[i]$ denotes the county of the $i^\text{th}$ household.
The model is:
\begin{align}
  p(\mu) & = \text{normal}(0, 1) \\
  p(\tau) & = \text{Uniform}(0, 100) \\
  p(\theta_j|\mu, \tau) & = \text{normal}(\mu, \tau^2) \\
  p(w_k) & = \text{normal}(0, 1) \\
  p(\sigma) & = \text{Uniform}(0, 100) \\
  p(\log y_i) & = \text{normal} ( {\bf w}^T {\bf x}_i + \theta_{j[i]}, \sigma^2).
\end{align}
We use FG-VI to approximate $p(\mu,\tau,\mathbf{\theta},\mathbf{w},\sigma|\mathbf{x},y)$. \\

\noindent
\textit{Stochastic volatility.} (n\!=\!103)
A time series model with 100 observations. The original model by \citet{Kim:1998} is given by:
%
\begin{align}
    p(\sigma) & = \text{Cauchy}^+(0, 2) \\
    p(\mu) & = \text{exponential}(1) \\
    p((\phi + 1) / 2) & = \text{Beta}(20, 1.5) \\
    p(z_i) & = \text{normal}(0, 1) \\
    h_1 & = \mu + \sigma z_1 / \sqrt{1 - \phi^2} \\
    h_{i > 1} & = \mu + \sigma z_i + \phi(h_{i - 1} - \mu) \\
    p(y_i \mid h_i) & = \text{normal}(0, \exp(h_i / 2)).
\end{align}
The original model was developed for a time series of 3000 observations, but we only work with the first 100 observations. For these observations, we use FG-VI to model $p(\sigma,\mu,\phi,\mathbf{z}|\mathbf{y})$.




\bibliography{references}

\end{document}